%% file: paper.tex
\documentclass[11pt]{article}
\usepackage{amsmath}
\usepackage{amsfonts}
\usepackage{booktabs}
\usepackage{amsmath}
\usepackage{enumitem}

\usepackage{times}
\usepackage{latexsym}
\usepackage{makecell}
\usepackage{times}
\usepackage{marvosym}
\usepackage{latexsym}
\usepackage{booktabs}
\usepackage{multirow}
\usepackage{graphicx} 
\usepackage{subcaption}
\usepackage{makecell}
\usepackage[normalem]{ulem}
\usepackage{amssymb}  
\usepackage{graphicx}
\usepackage{hyperref}
\usepackage{booktabs}
\usepackage{graphicx}
\usepackage{subcaption} 
\usepackage{amsmath}    

\usepackage[preprint]{acl}

\usepackage{times}
\usepackage{latexsym}

\usepackage[T1]{fontenc}

\usepackage[utf8]{inputenc}

\usepackage{microtype}

\usepackage{inconsolata}

\usepackage{graphicx}

\title{Neuron-Aware Active Few-Shot Learning for LLMs}

\author{
Zhuowei Chen \hspace{2pt} Liwei Chen \hspace{2pt} \\ \textbf{Christian Schunn} \hspace{2pt} \textbf{Raquel Coelho} \hspace{2pt} \textbf{Xiang Lorraine Li}\\
      University of Pittsburgh, USA\\
       \texttt{\{zhuowei.chen, xianglli\}@pitt.edu}
}

\begin{document}
\maketitle

\renewcommand{\arraystretch}{.7}

\input{0_abstract}
\input{0_introduction}

\input{1_related_work}

\input{2_method}

\input{3_experiments}
\input{4_conclusion}
\section*{Acknowledgement}
This research was supported in part by the University of Pittsburgh Center for Research Computing and Data, RRID:SCR\_022735, through the resources provided. Specifically, this work used the H2P cluster, which is supported by NSF award number OAC-2117681. The research is also funded by the Learning Research Development Center, University of Pittsburgh.

We are also grateful to the Pitt NLP group, and the anonymous reviewers for their constructive feedback and suggestions.

\input{appendix}

\bibliography{custom}

\end{document}

%% file: 0_abstract.tex
\begin{abstract} 

    Active Few-Shot Learning (AFSL) adapts LLMs to specialized domains by identifying the most valuable unlabeled samples for annotation and use as few-shot demonstrations, effectively reducing human annotation costs while promoting high performance. However, existing methods typically rely on output-level signals for the sample identification, such as predictive entropy or semantic similarities with test-time data based on external embeddings, which often overlook models' internal dynamics which could pinpoint specific knowledge gaps.
        To bridge this gap, we propose \textsc{\textbf{NeuFS}}, a \textbf{Neu}ron-Aware Active \textbf{F}ew-\textbf{S}hot Learning framework that shifts the selection paradigm from output-level proxies to models' internal dynamics. \textsc{NeuFS} utilizes neuron activation patterns to represent sample directly, and includes a dual-criteria selection strategy that: (1) ensures few-shot sample diversity with neuron patterns for broader example coverage, while (2) prioritizing on identifying informative and challenging few-shot samples LLMs tend to hallucinate by quantifying \textit{neuron consensus}.
    Experiments on three datasets demonstrate that \textsc{NeuFS} excels in both reasoning and text classification tasks, outperforming existing AFSL baselines.
    Ablation studies further highlight that internal neuron activations provide a more principled and effective selection signal than external embeddings, validating the superiority of the proposed \textsc{NeuFS}.

\end{abstract}



%% file: 0_introduction.tex
\section{Introduction}

    %

While recent advances in LLMs have greatly improved general tasks like text classification, these models often fail to perform well out-of-the-box in specialized domains such as education, medicine, or law. Adapting off-the-shelf models to these expert fields is typically made difficult by the scarcity of data and the computational costs required for fine-tuning.
To adapt LLMs for specific downstream tasks, few-shot in-context learning (ICL) has emerged as a popular training-free adaptation strategy. By providing examples within the query context, ICL has demonstrated promising performance in practical applications \citep{brown2020language}. The selection of few-shot has been identified as a critical factor for downstream task performance \citep{min2022rethinking}.  

Existing work on few-shot example selection typically assumes access to a large annotated dataset, where retrieving examples most similar to the test data at inference time yields the best performance~\cite{yu-etal-2023-retrieval, margatina2023active}. However, constructing large annotated datasets for test-time retrieval is oftentimes impractical, especially for specialized domains that require expensive expert-level annotations. Consequently, exploration of few-shot selection strategies under unlabeled data settings remains limited. Recent work has begun to address this gap by introducing active learning (AL) into few-shot ICL, i.e Active Few-Shot Learning (AFSL)~\cite{ahmadnia2025active}.

%
\begin{figure}[]
    \centering
    \includegraphics[width=1\linewidth]{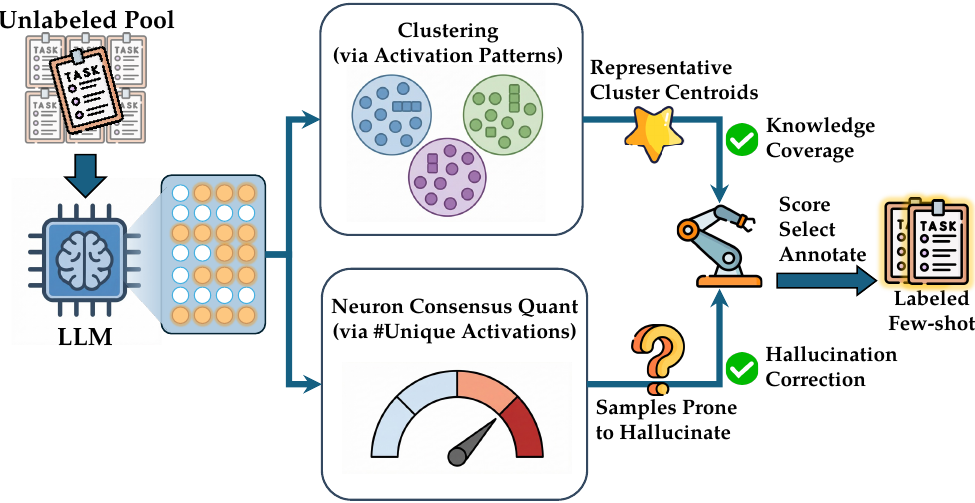}
    \caption{Neuron-Aware AFSL, which identifies informative and diverse few-shot samples for annotation. By leveraging activation patterns, \textsc{NeuFS} ensures broad knowledge coverage through activation pattern-based clustering while targeting informative, hallucination-prone samples via neuron consensus quantification.}
    \vspace{-1em}
    \label{fig:intro}
\end{figure}

AFSL extends traditional AL techniques to the few-shot paradigm, selecting the samples from a large unlabeled pool that are most likely to yield strong test-time performance, to be annotated as ICL demonstrations, providing a promising solution for reducing annotation costs while maintaining performance~\cite{xia2025selection, ahmadnia2025active}. 
AFSL methods largely follow the intuitions of mainstream AL approaches, which either prioritize highly uncertain samples to select informative samples \cite{meng2019weakly,gal2017deep}, ensure semantic diversity in the selected set \cite{yuan2020cold}, or consider both \cite{guo2024deuce,yu2023cold,hongjin2022selective}.

For informativeness, uncertainty measurement using entropy is commonly adopted. However, in the context of LLMs, it suffers from overconfidence and hallucination issues \cite{huang2025survey}, which LLMs can be confidently wrong. Moreover, the misaligned objectives of next token prediction and knowledge assessment further compound these limitations, meaning token predictive entropy cannot fully represent LLMs' intrinsic knowledge dynamics \cite{kuhn2023semantic}. 
For diversity, previous work relies on applying an external model to extract sample representations, then select samples through clustering \cite{ahmadnia2025active}, including KNN algorithms \cite{hongjin2022selective}. The strong assumption of semantic-knowledge equivalence causes the sub-optimal performance, that is, semantically similar samples can incentivize completely different knowledge when performing a specific task. Overall, most methods rely on output-level model signals \cite{guo2024deuce,ahmadnia2025active}.


To address these concerns, we shift from output-level signal toward internal model dynamics for AFSL.
Recent work indicates that neuron activation patterns in the Feed-Forward Networks (FFNs) of LLMs can link to multiple concrete knowledge concepts \cite{gao2024scaling}. Moreover, neuron activations could indicate potential hallucinations and model utility~\cite{chen2025llms, cao2025modelutilitylawevaluating}. Specifically, higher neuron consensus corresponds to fewer hallucinations. Thus, utilizing these internal dynamics provides a more robust and principled signal for AFSL by directly revealing the underlying activation patterns associated with the LLMs' internal knowledge and potential hallucinations.

To this end, we introduce \textsc{\textbf{NeuFS}}, a \textbf{Neuron-Aware Active Few-Shot Learning} framework that shifts the selection paradigm from output-level signals to internal model dynamics. As illustrated in Figure \ref{fig:intro}, \textsc{NeuFS} ensures broad knowledge-level coverage by diversifying neuron activation patterns via clustering and facilitates hallucination mitigation by leveraging neuron consensus as a uncertainty proxy. This strategy allows the selection to prioritize samples that trigger unique knowledge circuits or those where the LLM is prone to hallucinates. 
Extensive experiments on four models across three tasks ranging from complex reasoning to text classification demonstrate our method's strong generalizability and competitive 1$^{st}$ or 2$^{nd}$-ranking performance. Codes have been released\footnote{https://github.com/johnnychanv/NeuFS}. 



%% file: 1_related_work.tex
\section{Related Work}

\subsection{Active Learning}

AL aims at reducing annotation costs by selectively annotating the most valuable samples from a large pool of unlabeled data for model learning, while maximizing LLMs' performances \cite{cohn1994active}. 
For AL, previous work have proposed to select samples based on uncertainty \cite{meng2019weakly,gal2017deep}, semantic diversity \cite{yuan2020cold}, or jointly considers both \cite{guo2024deuce,yu2023cold,hongjin2022selective}.
As AL aligns with the nature of few-shot learning, where demonstrations are crucial as well, 
\citet{margatina2023active} analyzed AL in the few-shot setting, discovering that similarity between the testing query to the demonstrations is the key factor in few-shot selections, which aligns with their strong performances, such as TypiClust~\cite{hacohen2022active}.

However, most existing methods remain data-centric or rely on output-level signals, such as predictive entropy. In the context of LLMs, it suffers from overconfidence and hallucination issues \cite{huang2025survey}, and the misaligned objectives of language modeling and knowledge assessment further compound these limitations, meaning token predictive entropy cannot fully represent LLMs' intrinsic knowledge dynamics \cite{kuhn2023semantic}. 
Though gradient-based methods are being proposed recently~\cite{jung2025prismatic}, they often run with the ground truth label and require a significant computational resource overhead.
In contrast, our approach is model-centric. We bypass these lossy external proxies and output-level signals, and select demonstrations based directly on internal neuron activation patterns, ensuring that the selected samples align with the model’s cognitive processing.

\begin{figure*}[h!]
    \centering
    \includegraphics[width=\linewidth]{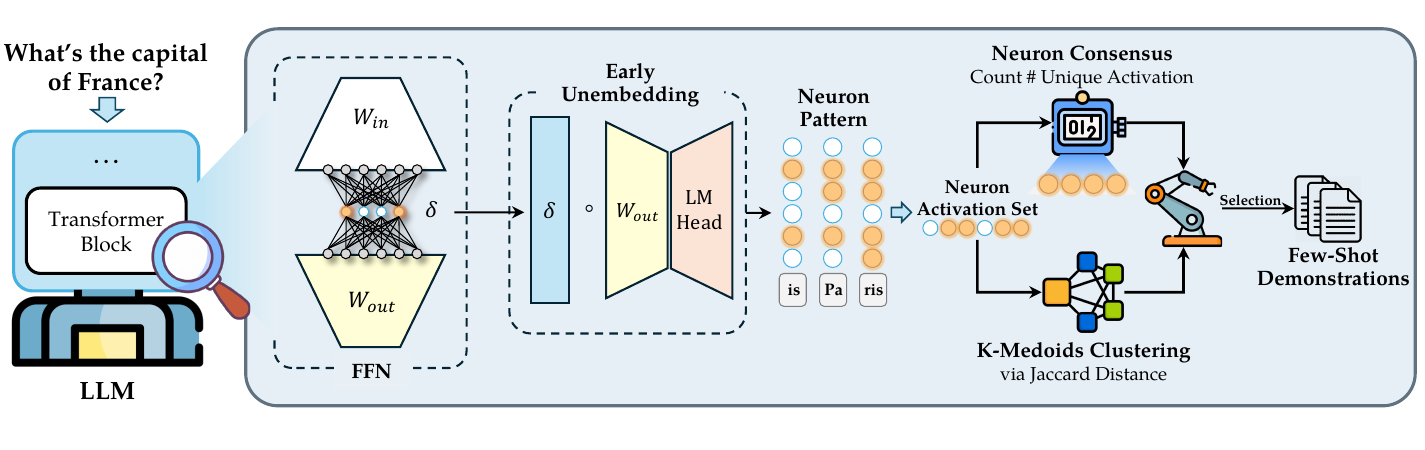}
    \caption{Overview of the proposed \textsc{NeuFS}. For each candidate, we extract the activation values from the FFN in each layer transformer during answer generation. Then we calculate the contribution score of each neuron via early unembedding, thereby identifying the activated neurons. Next, we perform Neuron Consensus Quantification by counting unique activations and K-Medoids Clustering via Jaccard distance for sample scoring and selection.}
    \label{fig:overview}

\vspace{-1.25em}
\end{figure*}

\subsection{Neuron Analysis}

Techniques for analyzing internal model states have evolved from linear probing~\citep{nostalgebraist2020interpreting} to sophisticated mechanistic interpretations~\citep{gao2024scaling}. 
Early work utilized the Logit Lens~\citep{nostalgebraist2020interpreting} and Tuned Lens~\citep{belrose2023eliciting} to map intermediate activation values in FFNs to vocabulary space. 
Recently, Sparse Autoencoders (SAEs) have achieved promising results in resolving neuron superposition and disentangling polysemantic features~\citep{gao2024scaling}.

Beyond feature extraction, recent studies have established a direct link between internal activation patterns and model reliability. 
\citet{chen2025llms} showed that internal neuron consensus, which is defined as the number of activated neurons across related neurons, serves as a strong indicator of hallucination, where higher consensus correlates with factual correctness. 
Similarly, \citet{cao2025modelutilitylawevaluating} introduced mechanism-interpretable metrics to evaluate model utility beyond surface-level performance, further validating the significance of internal signals.
However, while these insights link internal states to generation quality, their application to data selection remains underexplored. 
Our work bridges this gap by proposing a new paradigm which leverages these internal activation signals for AFSL.

%% file: 2_method.tex
\section{Methodology}

\textsc{NeuFS} operates as a pipeline for AFSL, instead of relying output-level signals, we proposed to shift the selection paradigm toward internal model dynamics by utilizing neuron activation patterns to represent samples directly. As illustrated in Figure \ref{fig:overview}, the framework consists of three sequential stages designed to capture the LLM's internal neuron states and uncertainty. First, we retrieve the raw activation values from the FFNs across all transformer layers for each candidate sample. Second, we employ \textbf{Neuron Activation Identification} via early unembedding to filter for neurons that contribute significantly to the model's final prediction. 
Afterwards, we execute \textbf{Neuron-Aware Active Few-Shot Selection}, which integrates \textit{Neuron-Aware Sample Diversification} with \textit{Neuron Consensus Quantification}. This combination prioritizes samples that trigger unique knowledge circuits that LLM tends to hallucinate, while ensuring the selected samples remain representative and diverse.

\subsection{Neuron Activation Identification}
\label{sec:neuron_ident}

To capture the internal dynamics of LLMs, we focus on neuron activations in FFNs, which are widely regarded as the key-value memories storing factual knowledge~\citep{geva-etal-2021-transformer}.
However, neuron activations does not strictly imply task relevance, neurons often exhibit polysemanticity or encode noise. 
To filter for neurons that genuinely support the model's internal reasoning process, we employ \textit{Early Unembedding}~\cite{chen2025llms}.

Formally, consider an LLM with $L$ layers. For an input sequence $x$, let $\mathbf{h}^{l}$ denote the input hidden state to the FFN at the $l$-th layer. The FFN consists of an up-projection matrix $\mathbf{W}_{\textit{in}}^{l} \in \mathbb{R}^{d \times d_{ff}}$, a non-linear activation function $\sigma(\cdot)$, and a down-projection matrix $\mathbf{W}_{\textit{out}}^{l} \in \mathbb{R}^{d_{ff} \times d}$, where $d_{ff}$ is the intermediate dimension of the FFN and $d$ represents the dimension of hidden states. The neuron activation values $\mathbf{k}^{l} \in \mathbb{R}^{d_{ff}}$ are computed as:
\begin{equation*}
    \mathbf{k}^{l} = \sigma(\mathbf{h}^{l} \mathbf{W}_{in}^{l}) ,
\end{equation*}
where $k^{l}$ represents the activation value at the transformer layer $l$, and $k_i^{l}$ is the $i$-th activation value.

While the activation value $k_i^{l}$ indicates how a neuron is activated, it does not directly reveal the semantic concept it promotes. To bridge the gap between internal states and output vocabulary, we analyze each neuron's contribution to the final prediction through the unembedding matrix in the LM head $\mathbf{E}_{u} \in \mathbb{R}^{d \times V}$, where $V$ is the vocabulary size, to decode neuron activations into a distribution over the vocabulary.

Let $\hat{y}$ be the prediction token generated by the model for the input $x$. The contribution score $S_{\hat{y}, i}^{l}$ of the $i$-th neuron in layer $l$ to the predicted token $\hat{y}$ is calculated by projecting its corresponding output vector directly into the vocabulary space:
\begin{equation*}
    S_{\hat{y}, i}^{l} = k_i^{l} \cdot (\mathbf{w}_{\textit{out}, i}^{l} \cdot \mathbf{e}_{\hat{y}}) ,
\end{equation*}
where $\mathbf{w}_{\textit{out}, i}^{l}$ is the $i$-th row of the down-projection matrix $\mathbf{W}_{\textit{out}}$, and $\mathbf{e}_{\hat{y}}$ is the embedding vector corresponding to the predicted token $\hat{y}$ in $\mathbf{E}_{u}$.

This score $S_{\hat{y}, i}^{l}$ explicitly quantifies how much neuron $i$ in layer $l$ contributes to the predicted token $\hat{y}$.
Following~\citet{chen2025llms}, we identify the set of \textit{valid} activated neurons $\mathcal{N}_{act}$ by filtering out those with contribution scores below the threshold. 
Specifically, the threshold is defined through a topk process, by selecting the highest k neuron activations across all activations in the FFN layers.
For a given threshold $\eta$, the activated neuron set is defined as the collection of neurons whose contribution scores exceed this value:
\begin{equation*}
    \mathcal{N}(x, \hat{y}) = \left\{ (i, l) \mid S_{\hat{y},i}^{l} > \eta \right\},
\end{equation*}
\vspace{-1.5em}
\begin{equation*}
    \mathcal{N}_{act}(x) = \mathcal{N}(x, {y_0}) \cup ... \cup \mathcal{N}(x, {y_t}),
\end{equation*}
where $y_t$ is the $t$-th token in the predicted sequence, and $\eta$ is dynamically determined to retain the top-$k$ most contributive neurons globally.
Specifically, for computational efficiency, we 
employ a two-stage filtering process.
We first retrieve the top-$n$\footnote{$n$ is set to 2000 for all experiments.} neurons from each layer based on raw activation to form a neuron pool. 
From this pool, we rank all neurons by their contribution scores and select the top-$k$ neurons as the activated neurons, which implicitly sets $\eta$ to the contribution score of the $k$-th ranked neuron.
This resulting set $\mathcal{N}_{\text{act}}$ serves as a sparse representation of the knowledge invoked by the LLM to process sample $x$, which forms the basis for subsequent selection.

\subsection{Neuron-Aware Active Few-Shot Selection}
\label{sec:select}

After identified the valid activated neurons for each candidate, we apply them to the selection process.
The core objective of \textsc{NeuFS} is to select a set of few-shot samples that are both \emph{representative} of the underlying task knowledge and \emph{informative} for correcting potential model hallucinations.
As shown in Figure \ref{fig:overview}, our pipeline integrates two distinct signals derived from the internal model dynamics: neuron activation patterns and neuron consensus.

Specifically, we first perform clustering based on the identified neuron sets to ensure the diversity of the selected examples.
Simultaneously, we quantify the neuron consensus for each sample by counting the unique activated neurons during the generation process, serving as a proxy for hallucination risk.
Finally, we employ a dual-criteria scoring that balances these two metrics to rank and select the optimal sample from each cluster.

\paragraph{Neuron-Aware Sample Diversification}
\label{sec:cluster}

To ensure the selected demonstrations cover diverse semantic and knowledge patterns, we group the unlabeled samples based on their internal activation patterns.
Unlike methods that rely on external embeddings, we utilize the sparse set of activated neurons $\mathcal{N}_{act}(x)$ as the sample representation.
Given that $\mathcal{N}_{act}(x)$ is a set of discrete indices (layer, neuron), we employ the Jaccard similarity to measure the similarity between two samples $x_i$ and $x_j$:
\begin{equation}
    D_{J}(x_i, x_j) = \frac{|\mathcal{N}_{act}(x_i) \cap \mathcal{N}_{act}(x_j)|}{|\mathcal{N}_{act}(x_i) \cup \mathcal{N}_{act}(x_j)|} ,
\end{equation}

We then apply the K-Medoids algorithm~\citep{kaufman1990partitioning} to partition the candidate pool into $C$ clusters, where $C$ corresponds to the target number of shots in few-shot experiments.

\paragraph{Neuron Consensus Quantification}
\label{sec:quant}

While diversification ensures coverage, it does not account for the quality or difficulty of the samples.
Drawing on findings that internal neuron activation patterns correlate strongly with model hallucinations~\cite{chen2025llms}, we introduce \textit{Neuron Consensus} as a metric for selecting informative and difficult samples.
Previous work suggests that higher neuron consensus, where a smaller number of unique neurons are activated during the generation process, corresponds to fewer hallucinations.
Conversely, samples with lower consensus indicate scenarios where the model's internal knowledge is sparse or conflicted, making them prone to hallucination.

For AFSL, these low-consensus samples are the most valuable as demonstration with human annotation guidance.
We quantify the neuron consensus $Q(x)$ simply as the count of unique valid activations identified in the early unembedding step, a higher value means lower consensus:
\begin{equation}
    Q(x) = |\mathcal{N}_{act}(x)| ,
\end{equation}
Our strategy prioritizes samples with higher $Q(x)$, aiming to retrieve samples that trigger more unique circuits which the LLMs tend to hallucinate.

\subsection{Sample Scoring and Selection}
\label{sec:scoring}

To further select the final demonstrations from each cluster, we design a dual-criteria scoring mechanism that integrates both representativeness (proximity to the cluster center) and informativeness (low neuron consensus).
For each cluster $\mathcal{C}_i$ with medoid $\mu_i$, we evaluate every sample $x \in \mathcal{C}_i$.
First, we normalize the distance and consensus metrics within the cluster to ensure comparable scales. Let $\tilde{D}_J(x, \mu_i)$ be the min-max normalized Jaccard distance to the medoid, and $\tilde{Q}(x)$ be the min-max normalized number of unique activated neurons.

We define the final selection score $Score(x)$ as a weighted combination of these normalized metrics:
\begin{equation}
    Score(x) = \tau \cdot \tilde{Q}(x) + (1 - \tau) \cdot \tilde{D}_J(x, \mu_i) .
\end{equation}
where $\tau \in [0, 1]$ is a hyperparameter controlling the trade-off between prioritizing low-consensus samples (hallucination correction) and representative samples (cluster centrality), which we have further discussions in Section.\ref{sec:hpms}.
The sample with the highest score in each cluster are selected.

%% file: 3_experiments.tex
\section{Experiments and Results}

\subsection{Setup}

We evaluate \textsc{NeuFS} against six baselines and their variants using four instruction-tuned LLMs (Llama3 3B/8B and Qwen3 4B/8B)~\cite{grattafiori2024llama,yang2025qwen3} on three reasoning and classification datasets: MMLU-Pro~\cite{wang2024mmlu}, Edu-Feedback~\cite{wu2023passive}, and TREC~\cite{li2002learning}, stats are in Table~\ref{tab:datastats}. We use non-reasoning mode with 5, 10, 20, and 30 shots, averaged over three inference runs with different seeds.

Moreover, we categorize methods by their {Information Type}, which denotes the source of information used for selection. These categories include \textit{Semantic} signals from external embedding models, \textit{Entropy} from output logits, \textit{Linguistic}\footnote{We extract sparse logical linguistic feature for feature augmentation, as demonstrated in Appendix.\ref{app:rst}}, and our proposed {neuron} activation patterns. Specifically, when combining different features, we compute a unified distance for the sample selection process by summing the distances derived from each feature.

\subsection{Datasets}
\label{sec:exp-dataset-descrip}
We evaluated \textsc{NeuFS} and baselines on three datasets from reasoning to classification tasks:

\begin{itemize}[itemsep=.15em]
    \item \textbf{MMLU-Pro} \cite{wang2024mmlu}. An advanced reasoning benchmark comprising 12,034 samples across 14 domains, which comprises ten options per question.
    \item \textbf{Edu-Feedback} \cite{wu2023passive}. A binary classification dataset for evaluating feedback quality (explanatory vs. non-explanatory) in educational systems. Collected from a real essay writing course, we split the annotated data into 1,799 training and 14,228 testing samples.
    \item \textbf{TREC} \cite{li2002learning}. A 6-way question classification dataset. It consists of short open-domain questions labeled into six classes, split into 5,452 training and 500 testing samples.
    
\end{itemize}

\subsection{Baselines}
\label{sec:algos}

We compare \textsc{NeuFS} against six AFSL baselines:
\begin{itemize}[itemsep=.15em]
    \item \textbf{Random.} Random selection with fixed seed.
    
    \item \textbf{Entropy.} Selects samples based on predictive uncertainty derived from output logits. We evaluate \textit{Highest-Entropy} (ranking samples by entropy score) and \textit{Diverse-Entropy} (stratified sampling across entropy bins).
    
    \item \textbf{TypiClust}~\cite{hacohen2022active}. A semantic-only approach that encodes text via BERT and selects representative samples with K-Means.
    
    \item \textbf{Patron}~\cite{yu2023cold}. Combines uncertainty and semantic. It applies graph-based uncertainty propagation to reduce entropy outliers before performing clustering selection.
    
    \item \textbf{FastVoteK}~\cite{hongjin2022selective}. A semantic diversity method that constructs a graph to select samples maximizing mutual distances.

    \item \textbf{VoteK}~\cite{hongjin2022selective}. Extension of FastVoteK, selects samples from diverse entropy bins to consider uncertainty.
    
\end{itemize}

\subsection{Results}
\label{sec:results}

By comparing methods across various \textbf{Info Types}, as detailed below, we show that shifting from output-level signals to internal neuron dynamics consistently yields better performance.

\input{results-MMLU-Pro}

\paragraph{Results on Reasoning Task.}
Table~\ref{tab:MMLU-res} highlights the limitations of traditional information types in complex reasoning scenarios, i.e MMLU-Pro. 
While methods relying on Semantic signals like TypiClust improve over random selection, they are consistently outperformed by \textsc{NeuFS}.
By utilizing the Neuron Info, \textsc{NeuFS} achieves the highest accuracy of {0.452} on Qwen3-4B and {0.418} on Qwen3-8B.
It is worth noting that \textsc{NeuFS} also surpasses Patron, a method combining both semantic and entropy signals. This result suggests that internal neuron consensus provides a more reliable signal for demonstration selection than combining external semantic density with predictive uncertainty.

\input{results-8B-TREC-EXP}
\input{results-4B-TREC-EXP}

\paragraph{Results on Classification Task.}

The advantage of the neuron-aware selection strategy consistency brings performance rises to classification tasks as shown in Table~\ref{tab:trec-edu-8b-res} and Table~\ref{tab:trec-edu-4b-res}. 

For 8B models shown in Table~\ref{tab:trec-edu-8b-res}, \textsc{NeuFS} consistently outperforms baselines that rely on entropy information. 
On the Edu-Feedback dataset, the proposed \textsc{NeuFS} outperforms all other baselines in terms of accuracy with both Llama3 and Qwen3 models, with competitive Macro-F1 values. Similarly, on the TREC dataset, the proposed NeuFS demonstrated state-of-the-art performances in both F1 values and accuracies with Qwen3-8B.

For 4B-level LLMs, this trend persists as presented in Table~\ref{tab:trec-edu-4b-res}. \textsc{NeuFS} demonstrates substantial gains in specialized domains. For instance, on the Edu-Feedback task with Qwen3-4B, it achieves a F1 of 0.692, significantly outperforming TypiClust and Highest-Entropy-based methods. While some baselines like Fast-VoteK (S.+L.) appear competitive on datasets like TREC, these methods often depend on the participation of external knowledge. \textsc{NeuFS}'s ability to achieve superior results using only internal neuron dynamics proves its efficacy as a more principled, model-aware selection strategy.

Rather than relying on surface-level cues such as semantic similarity or output probabilities, \textsc{NeuFS} directly leverages the model’s internal neuron activations to select informative examples for learning. It promotes diversity by identifying distinct activation patterns that capture complementary knowledge, and employs a neuron-consensus score to pinpoint instances where the model is most prone to errors or hallucinations. By explicitly targeting internal knowledge gaps, \textsc{NeuFS} improves sample efficiency and often outperforms prior selection strategies, particularly for domain-specific tasks.

\section{Ablation Study}


\subsection{Representation Variants}
\label{sec:ablation_rep}

To validate the effectiveness of using sparse neuron activations as sample representations, we compare \textsc{NeuFS} against various dense vector representations.
We compare with both \textit{Encoder-based} and \textit{Decoder-based} models.
For Encoder-based representations, we utilize SimCSE~\citep{gao-etal-2021-simcse}, a sentence embedding model based on BERT.
For Decoder-based representations, we utilizes the Qwen-Embedding-0.6B, which is specifically designed to generate high-quality text embeddings aligned with the Qwen series LLMs.

\input{ablations-rep-var}

As presented in Table~\ref{tab:ablation-reps}, results demonstrate that \textsc{NeuFS} consistently outperforms both dense representation variants.
While the Decoder-based Qwen generally surpasses the Encoder-based baseline SimCSE, it still underperforms compared to our neuron-aware method.
This confirms that the specific information provided by neuron activations, which directly links to knowledge circuits, is more effective for few-shot selection than the coarse-grained semantic proximity offered by dense vectors on reasoning and classification tasks.

\subsection{Impact of Neuron Sparsity $k$}
\label{sec:hpms}


The parameter $k$ determines the sparsity of the neuron representation $\mathcal{N}_{act}$. We vary $k$ from 2,000 to 10,000 to analyze how the granularity of the internal signal affects selection quality.

\begin{figure}[h!] 
    \centering
    \includegraphics[width=1\linewidth]{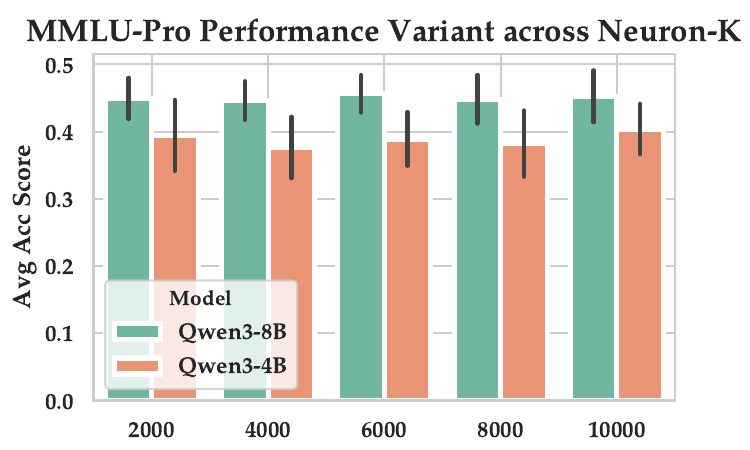}
    \caption{Ablation study on the $k$ which used to define the activation threshold. The bar plot shows average performance across 5,10,20,30 shots on Qwen3 series.}
    \label{fig:ablation_k}
    \vspace{-1.5em}
\end{figure}

As illustrated in Figure~\ref{fig:ablation_k}, the overall performance fluctuation is relatively mild, suggesting that \textsc{NeuFS} is generally robust to the specific activation threshold.
However, we observe a distinct difference in sensitivity between model sizes: the smaller Qwen3-4B model exhibits more noticeable performance variance compared to the Qwen3-8B, which remains highly stable across the spectrum.
This indicates that while larger models may possess redundant knowledge circuits that buffer against selection noise, smaller models are more sensitive to noises and the selection of $k$.

\subsection{Impact of Sample Scoring Weight \texorpdfstring{$\tau$}{tau}}
\label{sec:hpmtau}

\begin{figure}[h!] 
    \centering
    \begin{subfigure}[b]{\linewidth} 
        \centering
        \includegraphics[width=.9\linewidth]{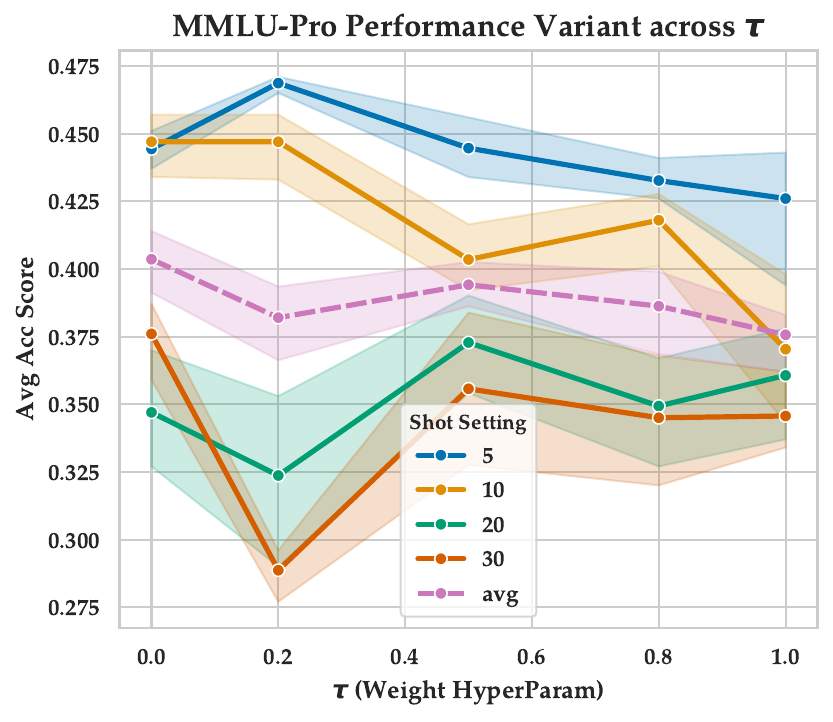}
        \caption{Performance on Qwen3 4B}
        \label{fig:tau_4b}
    \end{subfigure}
    \hfill 
    \begin{subfigure}[b]{\linewidth}
        \centering
        \includegraphics[width=.9\linewidth]{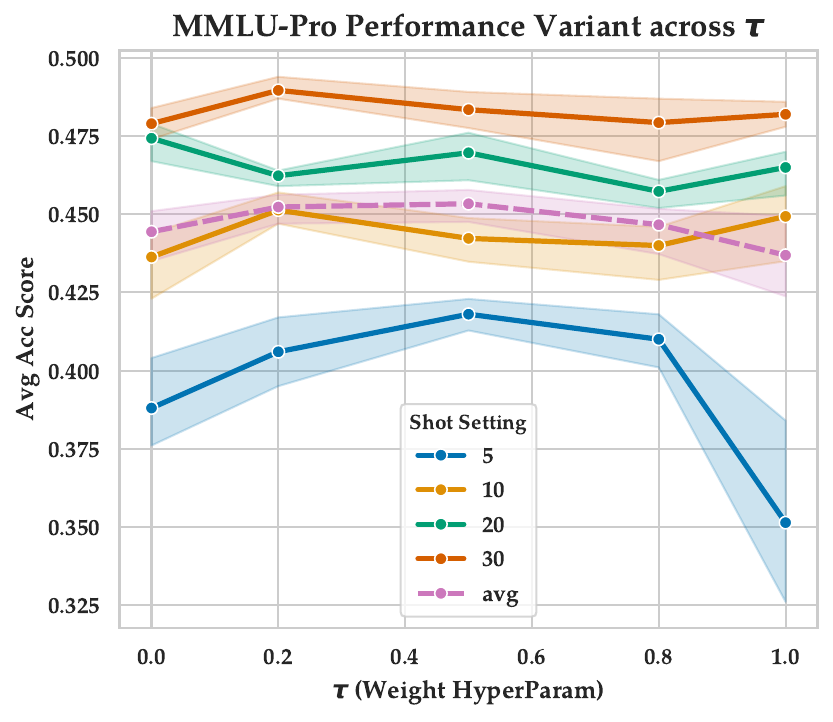}
        \caption{Performance on Qwen3 8B}
        \label{fig:tau_8b}
    \end{subfigure}
    
    \caption{Ablation study on the weight hyper-parameter $\tau$ across different model sizes. The shaded regions represent the standard deviation across three runs.}
    \label{fig:ablation_tau}
\end{figure}

The hyperparameter $\tau$ governs the trade-off between maximizing representativeness via cluster centric distance, which becomes dominant when $\tau$ approaches zero, and minimizing hallucination risk through Neuron Consensus, which prevails when $\tau$ approaches one. We analyze the impact of $\tau$ on model performance separately for different model sizes, as illustrated in Figure~\ref{fig:ablation_tau}.

\paragraph{For 4B-level models.}
As shown in Figure~\ref{fig:ablation_tau}(a), the smaller model exhibits significant volatility across different weight settings.
Crucially, though \textsc{NeuFS} achieved satisfying performance when $\tau$ is set to 0.5, the average performance represented by the dashed purple line achieves its maximum value when $\tau$ is set to 0, indicating that the 4B model could benefits most from pure sample diversification.
As $\tau$ increases, performance fluctuates and generally declines.
This suggests that neuron consensus signals could be noisier in smaller models. Less capacity leads to higher uncertainty across most candidates, reducing signal reliability and distinguishability. As a result, diversity and representativeness derived from neuron features become the dominant factors in AFSL.

\paragraph{For 8B-level models.}
In contrast, the larger model in Figure~\ref{fig:ablation_tau}(b) displays a much smoother trend.
The average performance clearly benefits from the integration of internal signals, peaking when $\tau$ is 0.5.
This indicates that for 8B models, which likely possess more stable and meaningful internal representations, the neuron consensus metric effectively complements diversity.
However, similar to the 4B model, pushing $\tau$ to 1 causes a sharp drop, confirming that while consensus is valuable, it cannot entirely replace diversity coverage.



%% file: results-MMLU-Pro.tex
\begin{table}[h!] 
  \centering
  \small 

\scalebox{1}{
  \begin{tabular}{clcc}
    \toprule
    \multicolumn{4}{c}{\textbf{(a) 8B Models}} \\ 
    \midrule
    Method & InfoType & Llama-3.1-8B & Qwen3-8B \\
    \midrule
    Random & N/A   & \textit{$ \text{0.325} _\text{±0.007}$} & $ \text{0.388} _\text{±0.079}$ \\
    Patron & S.+E. & $ \text{0.313} _\text{±0.013}$ & \textit{$ \text{0.416} _\text{±0.074}$} \\
    \midrule
    \multirow{2}[2]{*}{Entropy} & Highest & $ \text{0.317} _\text{±0.007}$ & $ \text{0.394} _\text{±0.095}$ \\
          & Diverse & $ \text{0.324} _\text{±0.003}$ & $ \text{0.392} _\text{±0.066}$ \\
    \midrule
    \multirow{3}[2]{*}{TypiClust} & S.    & $ \text{0.323} _\text{±0.007}$ & $ \text{0.398} _\text{±0.085}$ \\
          & L.    & $ \text{0.322} _\text{±0.006}$ & $ \text{0.382} _\text{±0.072}$ \\
          & S.+L. & $ \text{0.318} _\text{±0.006}$ & $ \text{0.383} _\text{±0.087}$ \\
    \midrule
    \multirow{3}[2]{*}{FastVoteK} & S.    & $ \text{0.316} _\text{±0.004}$ & $ \text{0.379} _\text{±0.103}$ \\
          & L.    & $ \text{0.323} _\text{±0.003}$ & $ \text{0.373} _\text{±0.089}$ \\
          & S.+L. & $ \text{0.318} _\text{±0.009}$ & $ \text{0.402} _\text{±0.075}$ \\
    \midrule
    \multirow{2}[2]{*}{VoteK} & S.+E. & $ \text{0.318} _\text{±0.004}$ & $ \text{0.391} _\text{±0.068}$ \\
          & S.+E.+L. & $ \text{0.309} _\text{±0.007}$ & $ \text{0.402} _\text{±0.075}$ \\
    \midrule
    NeuFS & Neuron & \textbf{$ \text{0.327} _\text{±0.007}$} & \textbf{$ \text{0.418} _\text{±0.069}$} \\
    \bottomrule
  \end{tabular}%

}
  
  \vspace{0.15cm} 

  \scalebox{1}{
  \begin{tabular}{clcc}
    \toprule
    \multicolumn{4}{c}{\textbf{(b) 3B \& 4B Models}} \\ 
    \midrule
    Method & InfoType & Llama-3.2-3B & Qwen3-4B \\
    \midrule
    Random & N/A   & \textit{$ \text{0.249} _\text{±0.005}$} & $ \text{0.412} _\text{±0.015}$ \\
    Patron & S.+E. & $ \text{0.244} _\text{±0.009}$ & $ \text{0.391} _\text{±0.045}$ \\
    \midrule
    \multirow{2}[2]{*}{Entropy} & Highest & $ \text{0.245} _\text{±0.005}$ & $ \text{0.430} _\text{±0.024}$ \\
          & Diverse & $ \text{0.244} _\text{±0.012}$ & $ \text{0.414} _\text{±0.019}$ \\
    \midrule
    \multirow{3}[2]{*}{TypiClust} & S.    & $ \text{0.242} _\text{±0.005}$ & \textit{$ \text{0.437} _\text{±0.016}$} \\
          & L.    & $ \text{0.248} _\text{±0.003}$ & $ \text{0.376} _\text{±0.026}$ \\
          & S.+L. & $ \text{0.243} _\text{±0.004}$ & $ \text{0.398} _\text{±0.028}$ \\
    \midrule
    \multirow{3}[2]{*}{FastVoteK} & S.    & $ \text{0.245} _\text{±0.001}$ & $ \text{0.386} _\text{±0.017}$ \\
          & L.    & $ \text{0.242} _\text{±0.002}$ & $ \text{0.401} _\text{±0.026}$ \\
          & S.+L. & $ \text{0.242} _\text{±0.003}$ & $ \text{0.420} _\text{±0.014}$ \\
    \midrule
    \multirow{2}[2]{*}{VoteK} & S.+E. & $ \text{0.246} _\text{±0.003}$ & $ \text{0.401} _\text{±0.012}$ \\
          & S.+E.+L. & $ \text{0.233} _\text{±0.008}$ & $ \text{0.374} _\text{±0.012}$ \\
    \midrule
    NeuFS & Neuron & \textbf{$ \text{0.251} _\text{±0.005}$} & \textbf{$ \text{0.452} _\text{±0.010}$} \\
    \bottomrule
  \end{tabular}%
  
  }

  \caption{Average accuracy ($\pm$ standard deviation) on the MMLU-Pro, aggregated across 5, 10, 20, and 30-shot settings. \textbf{Bold} and \textit{italics} indicate the best and second-best performance, respectively. InfoType denotes the source of information used for selection: Semantic (S.), Entropy (E.), Linguistic (L.), and Neuron activations.}
  \label{tab:MMLU-res}%
  \vspace{-1em}
\end{table}


%% file: results-8B-TREC-EXP.tex
\begin{table*}[htbp]
  \centering
  \scalebox{.75}{

    \begin{tabular}{clcccccccc}
    \toprule
    \multicolumn{2}{c}{Model} & \multicolumn{4}{c}{Llama-3.1-8B} & \multicolumn{4}{c}{Qwen3-8B} \\
    \midrule
    \multirow{2}[4]{*}{Method} & \multirow{2}[4]{*}{InfoType} & \multicolumn{2}{c}{Edu-Feedback} & \multicolumn{2}{c}{TREC} & \multicolumn{2}{c}{Edu-Feedback} & \multicolumn{2}{c}{TREC} \\
\cmidrule{3-10}          &       & Macro-F1 & Acc   & Macro-F1 & Acc   & Macro-F1 & Acc   & Macro-F1 & Acc \\
    \midrule
    Random & N/A   & $ \text{0.645} _\text{±0.053}$ & $ \text{0.662} _\text{±0.065}$ & $ \text{0.777} _\text{±0.052}$ & $ \text{0.764} _\text{±0.070}$ & $ \text{0.612} _\text{±0.084}$ & $ \text{0.625} _\text{±0.097}$ & $ \text{0.844} _\text{±0.022}$ & $ \text{0.823} _\text{±0.030}$ \\
    \midrule
    Patron & S.+E. & $ \text{0.643} _\text{±0.027}$ & $ \text{0.690} _\text{±0.025}$ & $ \text{0.807} _\text{±0.046}$ & $ \text{0.809} _\text{±0.045}$ & $ \text{0.659} _\text{±0.022}$ & \textit{$ \text{0.709} _\text{±0.037}$} & $ \text{0.825} _\text{±0.023}$ & $ \text{0.824} _\text{±0.017}$ \\
    \midrule
    \multirow{2}[2]{*}{Entropy} & Highest & $ \text{0.619} _\text{±0.027}$ & $ \text{0.625} _\text{±0.030}$ & $ \text{0.790} _\text{±0.029}$ & $ \text{0.785} _\text{±0.031}$ & $ \text{0.555} _\text{±0.049}$ & $ \text{0.557} _\text{±0.051}$ & $ \text{0.837} _\text{±0.035}$ & $ \text{0.839} _\text{±0.031}$ \\
          & Diverse & \textbf{$ \text{0.670} _\text{±0.028}$} & \textit{$ \text{0.696} _\text{±0.029}$} & $ \text{0.829} _\text{±0.030}$ & \textbf{$ \text{0.827} _\text{±0.033}$} & $ \text{0.632} _\text{±0.035}$ & $ \text{0.663} _\text{±0.070}$ & $ \text{0.856} _\text{±0.012}$ & $ \text{0.841} _\text{±0.017}$ \\
    \midrule
    \multirow{3}[2]{*}{TypiClust} & S.    & $ \text{0.642} _\text{±0.049}$ & $ \text{0.683} _\text{±0.071}$ & $ \text{0.798} _\text{±0.026}$ & $ \text{0.791} _\text{±0.035}$ & $ \text{0.605} _\text{±0.106}$ & $ \text{0.629} _\text{±0.122}$ & $ \text{0.831} _\text{±0.022}$ & $ \text{0.828} _\text{±0.022}$ \\
          & L.    & \textit{$ \text{0.664} _\text{±0.014}$} & $ \text{0.695} _\text{±0.021}$ & $ \text{0.806} _\text{±0.026}$ & $ \text{0.802} _\text{±0.038}$ & $ \text{0.626} _\text{±0.025}$ & $ \text{0.637} _\text{±0.033}$ & $ \text{0.835} _\text{±0.030}$ & $ \text{0.826} _\text{±0.040}$ \\
          & S.+L. & $ \text{0.623} _\text{±0.055}$ & $ \text{0.637} _\text{±0.066}$ & $ \text{0.814} _\text{±0.030}$ & $ \text{0.804} _\text{±0.039}$ & $ \text{0.589} _\text{±0.091}$ & $ \text{0.600} _\text{±0.101}$ & $ \text{0.851} _\text{±0.008}$ & $ \text{0.835} _\text{±0.005}$ \\
    \midrule
    \multirow{3}[2]{*}{FastVoteK} & S.    & $ \text{0.658} _\text{±0.012}$ & $ \text{0.694} _\text{±0.004}$ & $ \text{0.828} _\text{±0.024}$ & \textit{$ \text{0.825} _\text{±0.030}$} & $ \text{0.640} _\text{±0.027}$ & $ \text{0.662} _\text{±0.038}$ & $ \text{0.840} _\text{±0.017}$ & $ \text{0.831} _\text{±0.021}$ \\
          & L.    & $ \text{0.654} _\text{±0.027}$ & $ \text{0.676} _\text{±0.032}$ & \textit{$ \text{0.832} _\text{±0.046}$} & \textit{$ \text{0.825} _\text{±0.048}$} & $ \text{0.621} _\text{±0.059}$ & $ \text{0.632} _\text{±0.066}$ & $ \text{0.854} _\text{±0.018}$ & $ \text{0.843} _\text{±0.017}$ \\
          & S.+L. & $ \text{0.657} _\text{±0.042}$ & $ \text{0.688} _\text{±0.058}$ & $ \text{0.809} _\text{±0.051}$ & $ \text{0.813} _\text{±0.052}$ & $ \text{0.614} _\text{±0.097}$ & $ \text{0.631} _\text{±0.108}$ & \textit{$ \text{0.861} _\text{±0.022}$} & \textit{$ \text{0.854} _\text{±0.011}$} \\
    \midrule
    \multirow{2}[2]{*}{VoteK} & S.+E. & $ \text{0.660} _\text{±0.028}$ & $ \text{0.692} _\text{±0.027}$ & \textit{$ \text{0.832} _\text{±0.019}$} & $ \text{0.822} _\text{±0.020}$ & \textbf{$ \text{0.668} _\text{±0.020}$} & $ \text{0.701} _\text{±0.021}$ & $ \text{0.839} _\text{±0.006}$ & $ \text{0.839} _\text{±0.010}$ \\
          & S.+E.+L. & $ \text{0.653} _\text{±0.043}$ & $ \text{0.676} _\text{±0.054}$ & $ \text{0.818} _\text{±0.052}$ & $ \text{0.810} _\text{±0.051}$ & $ \text{0.631} _\text{±0.032}$ & $ \text{0.648} _\text{±0.047}$ & $ \text{0.839} _\text{±0.006}$ & $ \text{0.826} _\text{±0.022}$ \\
    \midrule
    NeuFS & Neuron & $ \text{0.660} _\text{±0.024}$ & \textbf{$ \text{0.698} _\text{±0.016}$} & \textbf{$ \text{0.834} _\text{±0.030}$} & $ \text{0.823} _\text{±0.036}$ & \textit{$ \text{0.663} _\text{±0.014}$} & \textbf{$ \text{0.711} _\text{±0.019}$} & \textbf{$ \text{0.862} _\text{±0.027}$} & \textbf{$ \text{0.858} _\text{±0.020}$} \\
    \bottomrule
    \end{tabular}%
  
  }
  \vspace{-.5em}
  \caption{Average performances ($\pm$ standard deviation) on the TREC and Edu-Feedback datasets with 8B models. \textbf{Bold} and \textit{italics} indicate the best and second-best performance, respectively.}
\vspace{-.5em}
  
  \label{tab:trec-edu-8b-res}%
\end{table*}%

%% file: results-4B-TREC-EXP.tex
\begin{table*}[h!]
  \centering
  \scalebox{.75}{

    \begin{tabular}{clcccccccc}
    \toprule
    \multicolumn{2}{c}{Model} & \multicolumn{4}{c}{Llama-3.2-3B} & \multicolumn{4}{c}{Qwen3-4B} \\
    \midrule
    \multirow{2}[4]{*}{Method} & \multirow{2}[4]{*}{InfoType} & \multicolumn{2}{c}{Edu-Feedback} & \multicolumn{2}{c}{TREC} & \multicolumn{2}{c}{Edu-Feedback} & \multicolumn{2}{c}{TREC} \\
\cmidrule{3-10}          &       & Macro-F1 & Acc   & Macro-F1 & Acc   & Macro-F1 & Acc   & Macro-F1 & Acc \\
    \midrule
    Random & N/A   & $ \text{0.557} _\text{±0.029}$ & $ \text{0.558} _\text{±0.030}$ & $ \text{0.699} _\text{±0.041}$ & $ \text{0.707} _\text{±0.049}$ & $ \text{0.653} _\text{±0.080}$ & $ \text{0.667} _\text{±0.094}$ & $ \text{0.859} _\text{±0.037}$ & $ \text{0.860} _\text{±0.037}$ \\
       \midrule
    Patron & S.+E. & $ \text{0.569} _\text{±0.111}$ & $ \text{0.597} _\text{±0.134}$ & $ \text{0.727} _\text{±0.051}$ & $ \text{0.749} _\text{±0.033}$ & \textit{$ \text{0.685} _\text{±0.027}$} & \textbf{$ \text{0.742} _\text{±0.030}$} & $ \text{0.842} _\text{±0.023}$ & $ \text{0.853} _\text{±0.024}$ \\
    \midrule
    \multirow{2}[2]{*}{Entropy} & Highest & $ \text{0.501} _\text{±0.080}$ & $ \text{0.506} _\text{±0.074}$ & $ \text{0.663} _\text{±0.022}$ & $ \text{0.638} _\text{±0.030}$ & $ \text{0.555} _\text{±0.065}$ & $ \text{0.557} _\text{±0.064}$ & $ \text{0.855} _\text{±0.012}$ & $ \text{0.857} _\text{±0.014}$ \\
          & Diverse & \textbf{$ \text{0.657} _\text{±0.034}$} & \textbf{$ \text{0.671} _\text{±0.042}$} & $ \text{0.736} _\text{±0.048}$ & $ \text{0.759} _\text{±0.020}$ & $ \text{0.671} _\text{±0.056}$ & $ \text{0.684} _\text{±0.066}$ & $ \text{0.862} _\text{±0.019}$ & $ \text{0.867} _\text{±0.013}$ \\
    \midrule
    \multirow{3}[2]{*}{TypiClust} & S.    & $ \text{0.545} _\text{±0.107}$ & $ \text{0.557} _\text{±0.116}$ & $ \text{0.741} _\text{±0.049}$ & $ \text{0.744} _\text{±0.039}$ & $ \text{0.626} _\text{±0.095}$ & $ \text{0.640} _\text{±0.111}$ & $ \text{0.853} _\text{±0.014}$ & $ \text{0.866} _\text{±0.003}$ \\
          & L.    & $ \text{0.606} _\text{±0.096}$ & $ \text{0.624} _\text{±0.116}$ & $ \text{0.712} _\text{±0.030}$ & $ \text{0.740} _\text{±0.069}$ & $ \text{0.636} _\text{±0.015}$ & $ \text{0.642} _\text{±0.017}$ & $ \text{0.862} _\text{±0.008}$ & $ \text{0.863} _\text{±0.008}$ \\
          & S.+L. & $ \text{0.537} _\text{±0.033}$ & $ \text{0.538} _\text{±0.033}$ & $ \text{0.753} _\text{±0.042}$ & $ \text{0.750} _\text{±0.049}$ & $ \text{0.613} _\text{±0.086}$ & $ \text{0.623} _\text{±0.096}$ & $ \text{0.855} _\text{±0.020}$ & $ \text{0.858} _\text{±0.013}$ \\
    \midrule
    \multirow{3}[2]{*}{FastVoteK} & S.    & $ \text{0.507} _\text{±0.046}$ & $ \text{0.509} _\text{±0.044}$ & \textbf{$ \text{0.760} _\text{±0.026}$} & $ \text{0.761} _\text{±0.030}$ & $ \text{0.673} _\text{±0.029}$ & $ \text{0.707} _\text{±0.047}$ & \textbf{$ \text{0.873} _\text{±0.010}$} & \textit{$ \text{0.871} _\text{±0.008}$} \\
          & L.    & $ \text{0.548} _\text{±0.104}$ & $ \text{0.555} _\text{±0.101}$ & $ \text{0.718} _\text{±0.044}$ & $ \text{0.739} _\text{±0.042}$ & $ \text{0.662} _\text{±0.046}$ & $ \text{0.673} _\text{±0.096}$ & $ \text{0.861} _\text{±0.027}$ & $ \text{0.867} _\text{±0.017}$ \\
          & S.+L. & $ \text{0.541} _\text{±0.059}$ & $ \text{0.543} _\text{±0.061}$ & $ \text{0.726} _\text{±0.034}$ & \textit{$ \text{0.775} _\text{±0.017}$} & $ \text{0.655} _\text{±0.083}$ & $ \text{0.679} _\text{±0.054}$ & $ \text{0.847} _\text{±0.026}$ & $ \text{0.857} _\text{±0.025}$ \\
    \midrule
    \multirow{2}[2]{*}{VoteK} & S.+E. & $ \text{0.526} _\text{±0.027}$ & $ \text{0.528} _\text{±0.029}$ & $ \text{0.755} _\text{±0.035}$ & $ \text{0.771} _\text{±0.023}$ & $ \text{0.664} _\text{±0.044}$ & $ \text{0.681} _\text{±0.056}$ & $ \text{0.853} _\text{±0.007}$ & $ \text{0.850} _\text{±0.012}$ \\
          & S.+E.+L. & $ \text{0.572} _\text{±0.065}$ & $ \text{0.577} _\text{±0.076}$ & \textit{$ \text{0.757} _\text{±0.025}$} & \textbf{$ \text{0.776} _\text{±0.024}$} & $ \text{0.681} _\text{±0.055}$ & $ \text{0.702} _\text{±0.067}$ & $ \text{0.854} _\text{±0.037}$ & $ \text{0.846} _\text{±0.023}$ \\
    \midrule
    NeuFS & Neuron & \textit{$ \text{0.624} _\text{±0.007}$} & \textit{$ \text{0.636} _\text{±0.008}$} & $ \text{0.754} _\text{±0.032}$ & $ \text{0.761} _\text{±0.044}$ & \textbf{$ \text{0.692} _\text{±0.021}$} & \textit{$ \text{0.725} _\text{±0.038}$} & \textit{$ \text{0.865} _\text{±0.019}$} & \textbf{$ \text{0.878} _\text{±0.015}$} \\
    \bottomrule
    \end{tabular}%
  
  }

\vspace{-.5em}

  \caption{Average performances ($\pm$ standard deviation) on the TREC and Edu-Feedback datasets with 3B and 4B models. \textbf{Bold} and \textit{italics} indicate the best and second-best performance, respectively.}

  \label{tab:trec-edu-4b-res}%

  \vspace{-.5em}
\end{table*}%

%% file: ablations-rep-var.tex
\begin{table}[h!]
  \centering
  \vspace{-0em}
  \scalebox{.75}{

    \begin{tabular}{cccc}
    \toprule
    \textbf{\#Shot} & \textbf{\makecell{NeuFS\\ w/ Qwen-Embed\footnote{https://huggingface.co/Qwen/Qwen3-Embedding-0.6B}}} & \textbf{\makecell{NeuFS\\ w/ simcse}} & \textbf{NeuFS} \\
    \midrule
    \multicolumn{4}{c}{MMLU-Pro} \\
    \midrule
    {5} & $ \text{0.2851} _\text{±0.0211}$ & \textbf{$ \text{0.3598} _\text{±0.0158}$} & $ \text{0.3341} _\text{±0.0232}$ \\
    {10} & $ \text{0.3823} _\text{±0.0113}$ & $ \text{0.3667} _\text{±0.0129}$ & \textbf{$ \text{0.3911} _\text{±0.0117}$} \\
    {20} & $ \text{0.4517} _\text{±0.0040}$ & \textbf{$ \text{0.4622} _\text{±0.0073}$} & $ \text{0.4593} _\text{±0.0054}$ \\
    {30} & $ \text{0.4863} _\text{±0.0039}$ & $ \text{0.4638} _\text{±0.0052}$ & \textbf{$ \text{0.4866} _\text{±0.0036}$} \\
    {Avg.} & $ \text{0.4013} _\text{±0.0887}$ & $ \text{0.4131} _\text{±0.0577}$ & \textbf{$ \text{0.4178} _\text{±0.0687}$} \\
    \midrule
    \multicolumn{4}{c}{Edu-Feedback} \\
    \midrule
    {5} & $ \text{0.6362} _\text{±0.0018}$ & $ \text{0.6308} _\text{±0.0009}$ & \textbf{$ \text{0.7088} _\text{±0.0009}$} \\
    {10} & $ \text{0.6492} _\text{±0.0018}$ & $ \text{0.6827} _\text{±0.0013}$ & \textbf{$ \text{0.6928} _\text{±0.0014}$} \\
    {20} & $ \text{0.5990} _\text{±0.0020}$ & \textbf{$ \text{0.7219} _\text{±0.0002}$} & $ \text{0.7033} _\text{±0.0012}$ \\
    {30} & $ \text{0.7162} _\text{±0.0005}$ & $ \text{0.7118} _\text{±0.0016}$ & \textbf{$ \text{0.7378} _\text{±0.0004}$} \\
    {Avg.} & $ \text{0.6502} _\text{±0.0489}$ & $ \text{0.6868} _\text{±0.0409}$ & \textbf{$ \text{0.7107} _\text{±0.0193}$} \\
    \midrule
    \multicolumn{4}{c}{TREC} \\
    \midrule
    {5} & \textbf{$ \text{0.8620} _\text{±0.0000}$} & $ \text{0.8433} _\text{±0.0031}$ & $ \text{0.8287} _\text{±0.0031}$ \\
    {10} & $ \text{0.8533} _\text{±0.0050}$ & $ \text{0.7847} _\text{±0.0012}$ & \textbf{$ \text{0.8727} _\text{±0.0012}$} \\
    {20} & $ \text{0.8027} _\text{±0.0023}$ & $ \text{0.7667} _\text{±0.0031}$ & \textbf{$ \text{0.8600} _\text{±0.0020}$} \\
    {30} & $ \text{0.8447} _\text{±0.0031}$ & $ \text{0.7847} _\text{±0.0031}$ & \textbf{$ \text{0.8687} _\text{±0.0023}$} \\
    {Avg.} & $ \text{0.8407} _\text{±0.0263}$ & $ \text{0.7948} _\text{±0.0334}$ & \textbf{$ \text{0.8575} _\text{±0.0199}$} \\
    \bottomrule
    \end{tabular}%
  }

\caption{Representation model variants.  }
\vspace{-1.5em}
  \label{tab:ablation-reps}%
\end{table}%

%% file: 4_conclusion.tex
\section{Conclusion}

In this work, we introduced \textsc{NeuFS}, an AFSL framework that shifts the selection paradigm from output-level signals to internal neuron dynamics.
By leveraging activation patterns, we proposed a dual-criteria scoring method that combines \textit{Neuron-Aware Sample Diversification} for coverage with \textit{Neuron Consensus} for hallucination mitigation.
Experiments on reasoning and classification tasks confirm \textsc{NeuFS}'s superiority over baselines, validating internal signals offer greater robustness than traditional similarity and output-level uncertainty metrics.
Beyond performance gains, this work highlights the critical role of model internal dynamics in data selection, encouraging future research to leverage internal model states for designing more robust and theoretically grounded strategies.

\section*{Limitations}
Despite the effectiveness of \textsc{NeuFS}, we acknowledge several limitations. The proposed framework relies on accessing internal FFN activations and unembedding matrices, limiting its applicability to open-weights models rather than black-box APIs. Also, while we employ two-stage filtering for efficiency, analyzing neuron patterns across large unlabeled pools incurs higher computational overhead compared to retrieving static pre-computed embeddings.

%% file: appendix.tex
\appendix

\section{Statistical Validation of Neuron Activation Signals}
\label{sec:appendix_statistical}

\subsection{RQ.1: Is Neuron Consensus Correlates Prediction Correctness?}
\label{sec:appendix_a1}

We revalidate the findings from prior work~\cite{cao2025modelutilitylawevaluating, chen2025llms} by measuring the relationship between prediction correctness and the number of unique neuron activations on MMLU-Pro. Specifically, we extract neuron activations by querying samples to LLMs individually, then record each sample's prediction correctness and its corresponding number of unique neuron activations (\#Unique Neuron Activations).

We group the correctly and incorrectly predicted samples into 50 bins each and compare the two distributions via a two-sample $t$-test, which confirms a statistically significant difference ($t = -3.5698$, $p < 0.001$): incorrectly predicted samples activate significantly more neurons than correctly predicted ones.

\begin{figure}[h]
    \centering
    \includegraphics[width=1\columnwidth]{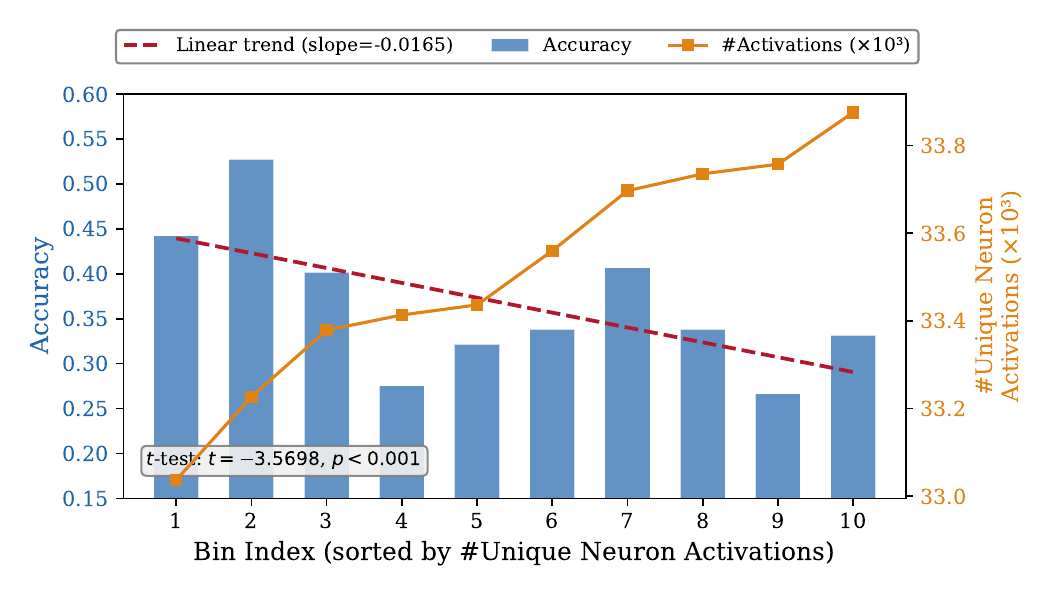}
    \caption{Relationship between \#Unique Neuron Activations and prediction accuracy on MMLU-Pro across 10 equal-width bins. Blue bars indicate accuracy per bin; the dashed red line shows the negative linear trend (slope $= -0.0165$). The orange line (right axis) shows the corresponding \#Unique Neuron Activations. A two-sample $t$-test over 50 bins confirms a significant difference between correct and incorrect predictions ($t = -3.5698$, $p < 0.001$).}
    \label{fig:a1_bin}
\end{figure}

To visualize the trend, we further aggregate all samples into 10 equal-width bins sorted by \#Unique Neuron Activations, as shown in Figure~\ref{fig:a1_bin}. The accuracy (blue bars) exhibits a general downward trend as the number of activated neurons increases (linear trend slope $= -0.0165$, red dashed line), while the orange line confirms the monotonically increasing activation count across bins. This validates that samples with lower neuron consensus tend to be more challenging for the model, supporting neuron consensus as an informativeness signal for identifying hallucination-prone samples.

\subsection{RQ.2: Does Low-Consensus Samples as Few-shot Improves Performance?}
\label{sec:appendix_a2}
 
Building on the finding that lower neuron consensus, i.e., higher \#neuron activations, indicates model uncertainty, we further investigate whether selecting demonstrations with lower neuron consensus as few-shot examples can improve inference performance. Specifically, we sample 30 queries from MMLU-Pro as inference inputs. For each query, we exclude the query itself from the dataset to construct a candidate pool for 5-shot selection.
 
For each inference query, we rank all candidates in the pool by their \#Unique Neuron Activations and use a sliding window (step size 1) to enumerate all possible 5-shot demonstration sets, where each set corresponds to a different level of neuron consensus among the selected demonstrations. We measure prediction accuracy across 20 equal-width bins. To eliminate query-specific difficulty bias, we compute a per-sample relative accuracy:
\begin{equation}
    \Delta\text{Acc} = \text{Acc}_{\text{sample}} - \overline{\text{Acc}}_{\text{query}},
\end{equation}
where $\text{Acc}_{\text{sample}}$ is the correctness of a given 5-shot configuration and $\overline{\text{Acc}}_{\text{query}}$ is the mean accuracy across all 5-shot pairs for the same query.
 
\begin{figure}[h]
    \centering
    \includegraphics[width=1\columnwidth]{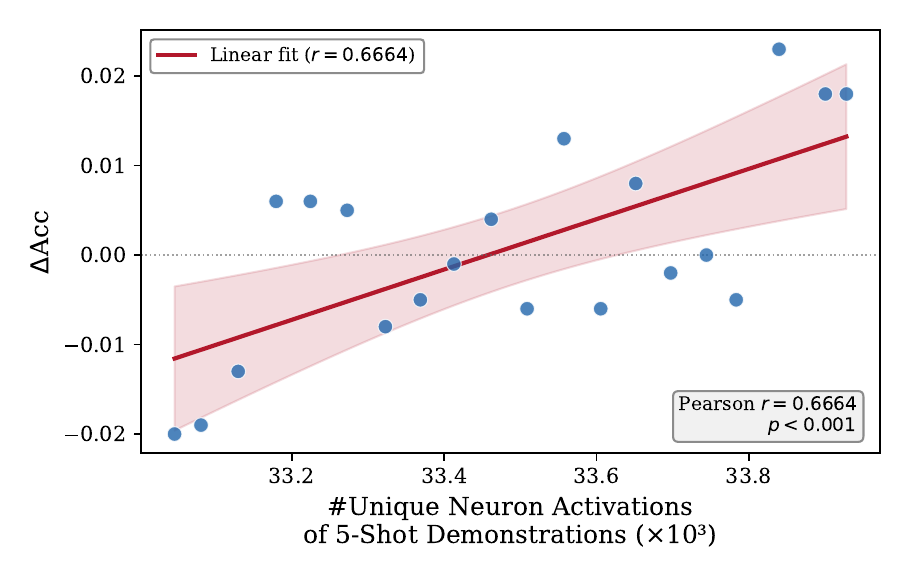}
    \caption{\#Unique Neuron Activations of 5-shot demonstrations vs.\ $\Delta$Acc on MMLU-Pro. Each point represents one of 20 equal-width bins. The red line shows the linear regression fit with a 95\% confidence band (shaded region). Pearson $r = 0.6664$, $p < 0.001$.}
    \label{fig:a2_scatter}
\end{figure}

Figure~\ref{fig:a2_scatter} presents a scatter plot of the 20 binned data points with a linear regression fit and 95\% confidence band. The strong positive correlation (Pearson $r = 0.6664$, $p < 0.001$) indicates that selecting demonstrations with higher \#Unique Neuron Activations --- i.e., samples the model is more uncertain about --- leads to measurably better inference performance.
 
\begin{figure}[h]
    \centering
    \includegraphics[width=1\columnwidth]{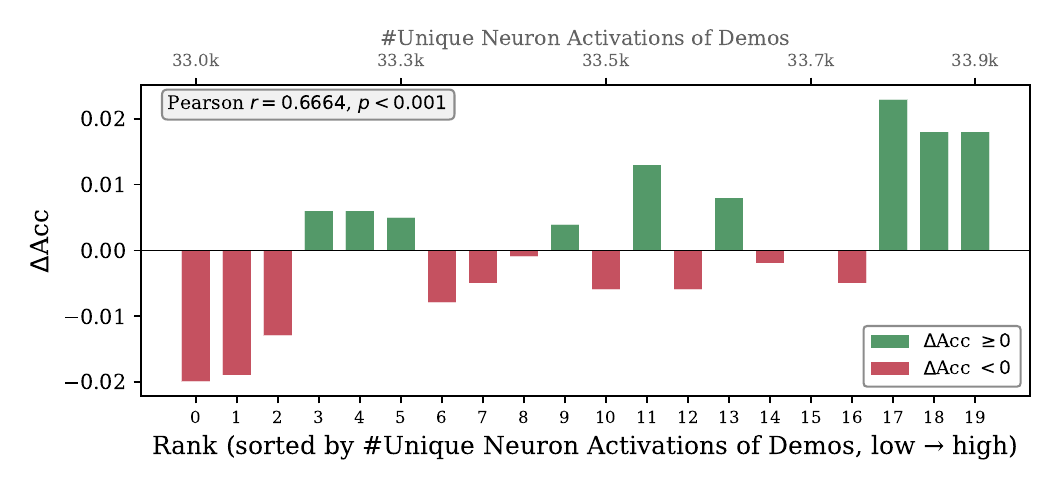}
    \caption{$\Delta$Acc across 20 rank bins sorted by \#Unique Neuron Activations of the 5-shot demonstrations (low $\to$ high). Green bars: $\Delta\text{Acc} \geq 0$; red bars: $\Delta\text{Acc} < 0$. The top axis shows the corresponding \#Unique Neuron Activations values. Demonstrations with lower neuron consensus (more unique activations) tend to improve inference performance.}
    \label{fig:a2_bar}
\end{figure}

Figure~\ref{fig:a2_bar} provides a complementary view, displaying $\Delta$Acc for each rank bin with color coding. The clear shift from predominantly negative $\Delta$Acc at low ranks (demonstrations with fewer unique activations) to predominantly positive $\Delta$Acc at high ranks (demonstrations with more unique activations) visually confirms the positive correlation: selecting challenging, high-consensus demonstrations as few-shot examples yields better model performance.
 
\section{Experiments}
\label{app:exp-results}

\subsection{Experiment Settings.} We experimented with  5, 10, 20, and 30 examples in few-shot setting and report the average performance. 
All experiments are done with the VLLM~\cite{kwon2023efficient} framework on four A100s, with a fixed temperature of 0.6. All reported results are averaged cross three runs, each with a different random seed.

\subsection{Baseline Algorithms}
\label{app:algos}

To perform comparison on various CSAL algorithms, we selected six baseline methods, as follows.
\begin{itemize}
    \item \textbf{Random.} We apply the random package in Python with a fix seed of 42 to perform 5, 10, 20, and 30 shot demonstration selection.
    \item \textbf{Entropy.} For entropy, we apply both highest-entropy and diverse-entropy selection strategies. We begin by extracting LLMs’ entropies using a few-shot prompting approach with 5 randomly sampled shots. The entropies are then computed from the output logits.
    For highest-entropy selection, following prior work~\cite{schroder2021revisiting}, we rank all samples by their entropy values and select the top 5, 10, 20, and 30 samples.
    For diverse-entropy selection, we partition the samples into multiple bins according to the number of shots, and then randomly select one sample from each bin to construct the final dataset.
    \item \textbf{TypiClust} \cite{hacohen2022active}. The method that purely considers the semantic information. Typiclust firstly encodes all textual samples with text representation models such as BERT, then performs unsupervised clustering with the classic KMeans algorithm, with the cluster number being set to the same as number of shots. Finally it selects sample from each cluster. 
    \item \textbf{Fast-Votek} \cite{hongjin2022selective}. This method is purely semantic-based, which performs a graph construction with semantic distances and the K nearest neighbour algorithm, then selects samples that could maximize the distances between all samples in the selected set.
    \item \textbf{Patron} \cite{yu2023cold}, which simultaneously considers both entropy information and semantic information. It constructed a graph based on semantic distances, then performed uncertainty propagation on the graph, which mitigates entropy outliers. Finally, it performs unsupervised clustering, selecting a sample from each cluster.
    \item \textbf{Votek} \cite{hongjin2022selective}, in addition to fast-votek, this algorithm incrementally considers the entropy information, selecting samples across different levels of entropy bins.

\end{itemize}

\subsection{Datasets}
\label{app:exp-dataset-descrip}
We used three datasets in our work, all data are anonymized.

\paragraph{Edu-Feedback.} Explanatory peer-feedback classification is a binary text classification dataset, which  distinguishes comments with explicit rationales from non-explanatory ones, serving as a key measure of review quality. It is a crucial task in building educational AI systems.

We collected and annotated more than 14,000 review samples from a real essay writing course. For experiments, we take 1799 samples as the training set, and the rest 14,228 samples as the testing set. 

\paragraph{TREC.} We use the TREC question classification dataset, coarse-grained, 6-way. It contains 5,452 training and 500 testing questions, each labeled with one of six categories: \textit{ABBR}, \textit{DESC}, \textit{ENTY}, \textit{HUM}, \textit{LOC}, and \textit{NUM}. Each instance is a short, open-domain question.

\paragraph{MMLU-Pro.} MMLU-Pro is an enhanced version of the Massive Multitask Language Understanding (MMLU) benchmark. It increases the difficulty by expanding the candidate options from 4 to 10 and focusing on more complex, reasoning-intensive questions across 14 diverse domains. This dataset provides a more robust evaluation of a model's advanced knowledge and analytical capabilities compared to the original benchmark. For experiments, we randomly selected 2000 samples as the candidate pool and the rest samples as the testing set.

\input{data_stat}

\subsection{Hyper-parameter Settings}
Here we provide the hyper-parameter setting for the hyperparameter $K$, which defines the threshold for activation neuron identification in Equation.3, and the hyperparameter $\tau$, which defines the weighting parameter in Equation.6.

\input{hpm-setting}

\subsection{Results}

We demonstrated the complete experimental results from Table.\ref{tab:llama-mmlu-pro} to Table.\ref{tab:qwen-trec}.

\input{LLama-full_results}

\input{Qwen-full_results}

\section{Linguistic Features.}
\label{app:rst}

\begin{figure}[h!]
    \centering
    \includegraphics[width=1\linewidth]{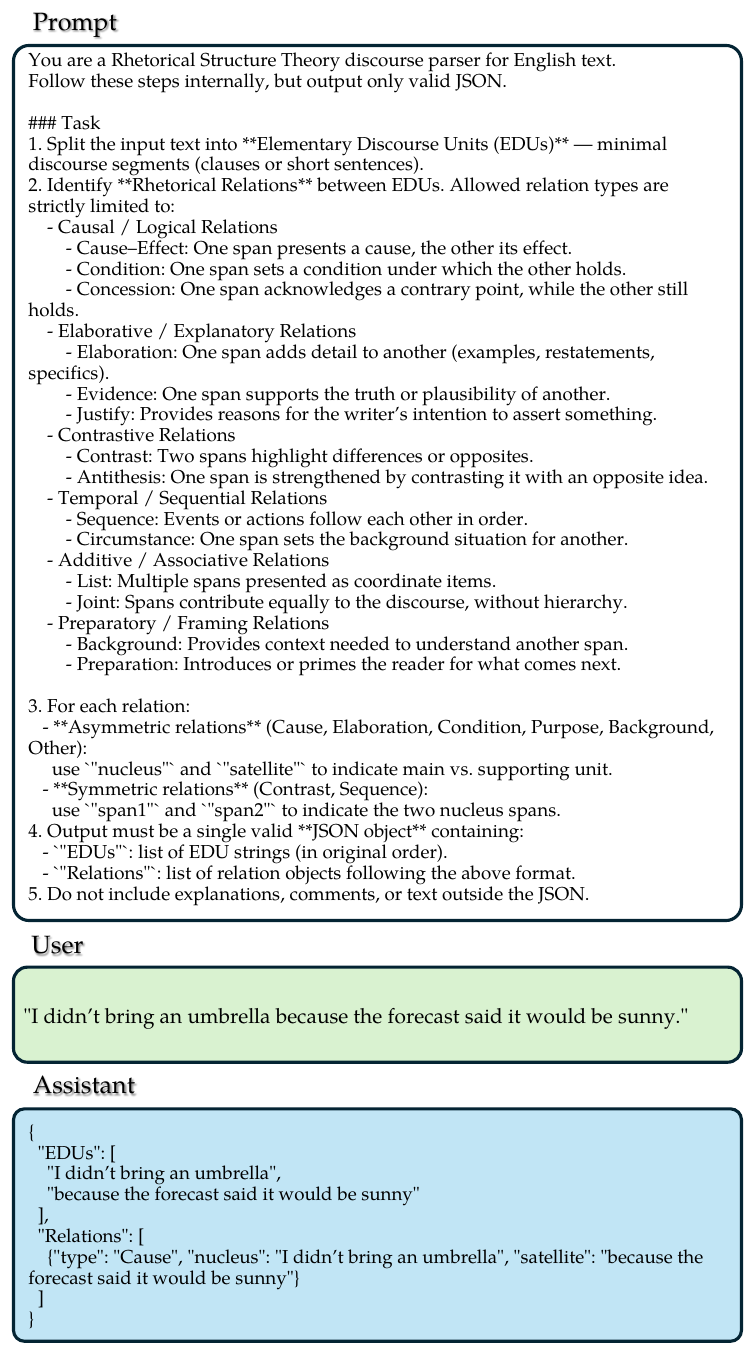}
    \caption{Example for RST Analysis.}
    \label{fig:rst}
\end{figure}

To incorporate additional linguistic information, we extract sparse linguistic features and integrate them into the selection process. We collect three types of linguistic features: count-based, Rhetorical Structure Theory (RST)~\cite{mann1988rhetorical} features and topical features. For count-based features, we count number of indicator features such as causal markers, rhetorical terms, relative clauses, critique and praise expressions, and action verbs.\footnote{NLTK (https://www.nltk.org) is employed for data processing, keyword dictionaries are manually defined.} For RST-based feature, we employed Qwen3-Next-80B~\cite{yang2025qwen3} to get the RST tree labels, then representing the results with a count-based vector. For RST Analysis, conversational example is demonstrated in Figure \ref{fig:rst}.

For topical analysis, we segmented long text into short sentences, clustered them with K-Means, and finally vectorized each sample using counts to obtain sparse features.

%% file: data_stat.tex
\begin{table}[h]
  \centering
        \scalebox{.70}{
            \begin{tabular}{ccccc}
    \toprule
    \textbf{Dataset} & \textbf{Source Domain} & \textbf{\#Class} & \textbf{\#Candidate} & \textbf{\#Test} \\
    \midrule
    MMLU-Pro & Multi-disciplinary  & 10     & 2000 & 10032 \\
    Edu-Feedback & Peer-feedback & 2     & 1799  & 14228 \\
    TREC  & Question & 6     & 5452  & 500 \\
    \bottomrule
    \end{tabular}%
        }
      \caption{Data Statistics.}
  \label{tab:datastats}%
\end{table}%

%% file: hpm-setting.tex
\begin{table}[h!]
  \centering
    \begin{tabular}{ccc}
    \toprule
    Model & K     & $\tau$ \\
    \midrule
    \multicolumn{3}{c}{MMLU-Pro} \\
    \midrule
    Llama-3.1-8B & 6000  & 0.8 \\
    Llama-3.2-3B & 6000  & 1 \\
    Qwen3-8B & 8000  & 0.5 \\
    Qwen3-4B-Instruct-2507 & 8000  & 0.5 \\
    \midrule
    \multicolumn{3}{c}{Edu-Feedback} \\
    \midrule
    Llama-3.1-8B & 500   & 0.2 \\
    Llama-3.2-3B & 1000  & 1 \\
    Qwen3-8B & 1000  & 0.5 \\
    Qwen3-4B-Instruct-2507 & 4000  & 0.5 \\
    \midrule
    \multicolumn{3}{c}{TREC} \\
    \midrule
    Llama-3.1-8B & 3000  & 0.25 \\
    Llama-3.2-3B & 4000  & 0.5 \\
    Qwen3-8B & 5000  & 1 \\
    Qwen3-4B-Instruct-2507 & 5000  & 1 \\
    \bottomrule
    \end{tabular}%
  \caption{Hyper-parameter Setting}
  \label{tab:addlabel}%
\end{table}%

%% file: LLama-full_results.tex
\begin{table*}[htbp]
  \centering

    \scalebox{.5}{
        \begin{tabular}{cccccccccccc}
    \toprule
    \multicolumn{12}{c}{Llama-3.1-8B-Instruct} \\
    \midrule
    \multirow{2}[4]{*}{Method} & \#Shot & \multicolumn{2}{c}{5} & \multicolumn{2}{c}{10} & \multicolumn{2}{c}{20} & \multicolumn{2}{c}{30} & \multicolumn{2}{c}{Avg.} \\
\cmidrule{2-12}          & InfoType & Macro-F1 & Acc   & Macro-F1 & Acc   & Macro-F1 & Acc   & Macro-F1 & Acc   & Macro-F1 & Acc \\
    \midrule
    Random & N/A   & 0.3401±0.0142 & 0.3292±0.0193 & 0.3428±0.0179 & 0.3318±0.0231 & 0.3309±0.0277 & 0.3208±0.0347 & 0.3270±0.0235 & 0.3165±0.0291 & 0.3352±0.0075 & 0.3246±0.0072 \\
    \midrule
    Patron & S.+E. & 0.3418±0.0117 & 0.3305±0.0177 & 0.3275±0.0213 & 0.3157±0.0276 & 0.3135±0.0317 & 0.3020±0.0362 & 0.3135±0.0336 & 0.3055±0.0392 & 0.3241±0.0135 & 0.3134±0.0128 \\
    \midrule
    \multirow{2}[2]{*}{Entropy} & Highest & 0.3358±0.0088 & 0.3236±0.0139 & 0.3297±0.0280 & 0.3205±0.0325 & 0.3242±0.0317 & 0.3163±0.0366 & 0.3147±0.0322 & 0.3066±0.0354 & 0.3261±0.0090 & 0.3167±0.0074 \\
          & Diverse & 0.3379±0.0183 & 0.3266±0.0252 & 0.3302±0.0269 & 0.3194±0.0334 & 0.3355±0.0261 & 0.3248±0.0334 & 0.3354±0.0190 & 0.3243±0.0252 & 0.3348±0.0033 & 0.3237±0.0031 \\
    \midrule
    \multirow{3}[2]{*}{TypiClust} & S.    & 0.3411±0.0199 & 0.3321±0.0258 & 0.3371±0.0196 & 0.3262±0.0252 & 0.3281±0.0255 & 0.3174±0.0311 & 0.3282±0.0226 & 0.3172±0.0302 & 0.3336±0.0065 & 0.3232±0.0073 \\
          & L.    & 0.3387±0.0201 & 0.3293±0.0258 & 0.3344±0.0248 & 0.3247±0.0292 & 0.3291±0.0290 & 0.3188±0.0338 & 0.3273±0.0265 & 0.3165±0.0335 & 0.3324±0.0052 & 0.3223±0.0058 \\
          & S.+L. & 0.3350±0.0183 & 0.3260±0.0243 & 0.3250±0.0259 & 0.3145±0.0307 & 0.3249±0.0296 & 0.3150±0.0332 & 0.3248±0.0285 & 0.3145±0.0347 & 0.3274±0.0050 & 0.3175±0.0057 \\
    \midrule
    \multirow{3}[2]{*}{fast-votek} & S.    & 0.3307±0.0201 & 0.3211±0.0246 & 0.3234±0.0306 & 0.3137±0.0336 & 0.3247±0.0280 & 0.3160±0.0337 & 0.3222±0.0280 & 0.3134±0.0329 & 0.3252±0.0038 & 0.3160±0.0035 \\
          & L.    & 0.3349±0.0226 & 0.3268±0.0280 & 0.3318±0.0287 & 0.3216±0.0338 & 0.3305±0.0244 & 0.3209±0.0292 & 0.3331±0.0238 & 0.3238±0.0294 & 0.3326±0.0019 & 0.3233±0.0026 \\
          & S.+L. & 0.3367±0.0229 & 0.3281±0.0277 & 0.3332±0.0260 & 0.3235±0.0295 & 0.3209±0.0361 & 0.3123±0.0384 & 0.3190±0.0301 & 0.3086±0.0339 & 0.3275±0.0088 & 0.3181±0.0092 \\
    \midrule
    \multirow{2}[2]{*}{votek} & S.+E. & 0.3308±0.0223 & 0.3210±0.0266 & 0.3261±0.0265 & 0.3161±0.0304 & 0.3312±0.0278 & 0.3212±0.0322 & 0.3229±0.0302 & 0.3128±0.0359 & 0.3277±0.0040 & 0.3178±0.0041 \\
          & S.+E.+L. & 0.3232±0.0272 & 0.3142±0.0289 & 0.3258±0.0263 & 0.3145±0.0301 & 0.3139±0.0294 & 0.3021±0.0331 & 0.3137±0.0333 & 0.3039±0.0387 & 0.3191±0.0063 & 0.3087±0.0066 \\
    \midrule
    NeuronPattern & Neuron & 0.3463±0.0124 & 0.3371±0.0174 & 0.3374±0.0144 & 0.3263±0.0182 & 0.3321±0.0227 & 0.3224±0.0292 & 0.3320±0.0232 & 0.3229±0.0288 & 0.3369±0.0067 & 0.3272±0.0068 \\
    \midrule
    \multicolumn{12}{c}{Llama-3.2-3B-Instruct} \\
    \midrule
    \multirow{2}[4]{*}{Method} & \#Shot & \multicolumn{2}{c}{5} & \multicolumn{2}{c}{10} & \multicolumn{2}{c}{20} & \multicolumn{2}{c}{30} & \multicolumn{2}{c}{Avg.} \\
\cmidrule{2-12}          & InfoType & Macro-F1 & Acc   & Macro-F1 & Acc   & Macro-F1 & Acc   & Macro-F1 & Acc   & Macro-F1 & Acc \\
    \midrule
    Random & N/A   & 0.2349±0.0579 & 0.2481±0.0470 & 0.2401±0.0612 & 0.2532±0.0499 & 0.2267±0.0728 & 0.2422±0.0566 & 0.2433±0.0588 & 0.2521±0.0479 & 0.2362±0.0072 & 0.2489±0.0050 \\
    \midrule
    Patron & S.+E. & 0.2374±0.0601 & 0.2516±0.0485 & 0.2339±0.0553 & 0.2429±0.0411 & 0.2188±0.0725 & 0.2321±0.0564 & 0.2367±0.0702 & 0.2494±0.0549 & 0.2317±0.0087 & 0.2440±0.0087 \\
    \midrule
    \multirow{2}[2]{*}{Entropy} & Highest & 0.2343±0.0544 & 0.2484±0.0444 & 0.2361±0.0634 & 0.2501±0.0507 & 0.2239±0.0673 & 0.2388±0.0517 & 0.2276±0.0661 & 0.2417±0.0525 & 0.2305±0.0057 & 0.2447±0.0054 \\
          & Diverse & 0.2003±0.0796 & 0.2261±0.0570 & 0.2285±0.0716 & 0.2469±0.0568 & 0.2364±0.0667 & 0.2507±0.0539 & 0.2392±0.0621 & 0.2501±0.0496 & 0.2261±0.0178 & 0.2435±0.0117 \\
    \midrule
    \multirow{3}[2]{*}{TypiClust} & S.    & 0.2420±0.0520 & 0.2492±0.0405 & 0.2313±0.0649 & 0.2429±0.0495 & 0.2260±0.0596 & 0.2371±0.0465 & 0.2307±0.0557 & 0.2396±0.0445 & 0.2325±0.0068 & 0.2422±0.0052 \\
          & L.    & 0.2389±0.0509 & 0.2477±0.0420 & 0.2411±0.0543 & 0.2517±0.0445 & 0.2299±0.0698 & 0.2434±0.0537 & 0.2377±0.0632 & 0.2474±0.0514 & 0.2369±0.0049 & 0.2475±0.0034 \\
          & S.+L. & 0.2409±0.0539 & 0.2483±0.0422 & 0.2276±0.0624 & 0.2379±0.0458 & 0.2294±0.0731 & 0.2427±0.0562 & 0.2328±0.0613 & 0.2421±0.0476 & 0.2327±0.0059 & 0.2428±0.0043 \\
    \midrule
    \multirow{3}[2]{*}{fast-votek} & S.    & 0.2342±0.0552 & 0.2457±0.0440 & 0.2333±0.0578 & 0.2432±0.0454 & 0.2331±0.0610 & 0.2447±0.0488 & 0.2348±0.0578 & 0.2452±0.0468 & 0.2338±0.0008 & 0.2447±0.0011 \\
          & L.    & 0.2215±0.0806 & 0.2394±0.0592 & 0.2257±0.0792 & 0.2429±0.0600 & 0.2298±0.0720 & 0.2431±0.0565 & 0.2322±0.0675 & 0.2440±0.0531 & 0.2273±0.0047 & 0.2423±0.0020 \\
          & S.+L. & 0.2237±0.0716 & 0.2379±0.0519 & 0.2313±0.0711 & 0.2458±0.0553 & 0.2249±0.0755 & 0.2408±0.0559 & 0.2344±0.0620 & 0.2431±0.0497 & 0.2286±0.0051 & 0.2419±0.0033 \\
    \midrule
    \multirow{2}[2]{*}{votek} & S.+E. & 0.2321±0.0503 & 0.2411±0.0391 & 0.2342±0.0609 & 0.2465±0.0486 & 0.2326±0.0714 & 0.2458±0.0562 & 0.2386±0.0616 & 0.2492±0.0504 & 0.2344±0.0029 & 0.2456±0.0034 \\
          & S.+E.+L. & 0.2081±0.0646 & 0.2227±0.0482 & 0.2140±0.0771 & 0.2314±0.0568 & 0.2246±0.0731 & 0.2380±0.0558 & 0.2250±0.0721 & 0.2398±0.0559 & 0.2179±0.0083 & 0.2330±0.0078 \\
    \midrule
    NeuronPattern & Neuron & 0.2435±0.0512 & 0.2561±0.0439 & 0.2434±0.0534 & 0.2534±0.0445 & 0.2393±0.0571 & 0.2499±0.0466 & 0.2333±0.0589 & 0.2447±0.0478 & 0.2399±0.0048 & 0.2510±0.0049 \\
    \bottomrule
    \end{tabular}%

    }

  \caption{Experimental results with Llama series on MMLU-Pro.}

  \label{tab:llama-mmlu-pro}%
\end{table*}%

\begin{table*}[htbp]
  \centering

\scalebox{.5}{

    \begin{tabular}{cccccccccccc}
    \toprule
    \multicolumn{12}{c}{Llama-3.1-8B-Instruct} \\
    \midrule
    \multirow{2}[3]{*}{Method} & \#Shot & \multicolumn{2}{c}{5} & \multicolumn{2}{c}{10} & \multicolumn{2}{c}{20} & \multicolumn{2}{c}{30} & \multicolumn{2}{c}{Avg.} \\
\cmidrule{2-12}          & InfoType & Macro-F1 & Acc   & Macro-F1 & Acc   & Macro-F1 & Acc   & Macro-F1 & Acc   & Macro-F1 & Acc \\
        \midrule

    Random & N/A   & 0.6424±0.0048 & 0.6581±0.0074 & 0.5728±0.0101 & 0.5739±0.0108 & 0.6691±0.0042 & 0.6899±0.0072 & 0.6974±0.0037 & 0.7273±0.0066 & 0.6454±0.0534 & 0.6623±0.0654 \\
        \midrule

    Patron & S.+E. & 0.6636±0.0010 & 0.7263±0.0041 & 0.6046±0.0036 & 0.6862±0.0032 & 0.6565±0.0048 & 0.6751±0.0078 & 0.6488±0.0047 & 0.6703±0.0080 & 0.6434±0.0266 & 0.6895±0.0254 \\
    \midrule
    \multirow{2}[1]{*}{Entropy} & Highest & 0.6066±0.0076 & 0.6128±0.0091 & 0.5878±0.0104 & 0.5897±0.0114 & 0.6314±0.0080 & 0.6384±0.0096 & 0.6497±0.0094 & 0.6571±0.0111 & 0.6189±0.0272 & 0.6245±0.0295 \\
          & Diverse & 0.6447±0.0055 & 0.6643±0.0084 & 0.6468±0.0020 & 0.6775±0.0055 & 0.6907±0.0052 & 0.7183±0.0107 & 0.6967±0.0065 & 0.7223±0.0102 & 0.6697±0.0278 & 0.6956±0.0291 \\
        \midrule
    \multirow{3}[0]{*}{TypiClust} & S.    & 0.5720±0.0098 & 0.5770±0.0117 & 0.6462±0.0036 & 0.7066±0.0043 & 0.6649±0.0047 & 0.7265±0.0021 & 0.6828±0.0006 & 0.7202±0.0065 & 0.6415±0.0487 & 0.6826±0.0709 \\
          & L.    & 0.6608±0.0025 & 0.7153±0.0040 & 0.6575±0.0012 & 0.6987±0.0045 & 0.6527±0.0101 & 0.6649±0.0127 & 0.6836±0.0061 & 0.6991±0.0081 & 0.6636±0.0137 & 0.6945±0.0212 \\
          & S.+L. & 0.5600±0.0063 & 0.5618±0.0072 & 0.6000±0.0104 & 0.6052±0.0123 & 0.6466±0.0036 & 0.6740±0.0073 & 0.6857±0.0034 & 0.7082±0.0065 & 0.6231±0.0547 & 0.6373±0.0661 \\
        \midrule
    \multirow{3}[1]{*}{fast-votek} & S.    & 0.6463±0.0002 & 0.6965±0.0061 & 0.6551±0.0014 & 0.6889±0.0052 & 0.6568±0.0020 & 0.6956±0.0070 & 0.6749±0.0039 & 0.6951±0.0066 & 0.6583±0.0120 & 0.6940±0.0035 \\
          & L.    & 0.6179±0.0030 & 0.6435±0.0069 & 0.6722±0.0005 & 0.7124±0.0044 & 0.6751±0.0064 & 0.6922±0.0088 & 0.6489±0.0077 & 0.6559±0.0089 & 0.6535±0.0265 & 0.6760±0.0319 \\
          & S.+L. & 0.5953±0.0068 & 0.6048±0.0089 & 0.6775±0.0008 & 0.7333±0.0044 & 0.6646±0.0041 & 0.6911±0.0074 & 0.6900±0.0012 & 0.7232±0.0043 & 0.6568±0.0423 & 0.6881±0.0584 \\
    \midrule
    \multirow{2}[1]{*}{votek} & S.+E. & 0.6480±0.0019 & 0.7030±0.0036 & 0.6402±0.0038 & 0.6710±0.0075 & 0.6481±0.0047 & 0.6673±0.0074 & 0.7017±0.0036 & 0.7248±0.0064 & 0.6595±0.0284 & 0.6915±0.0274 \\
          & S.+E.+L. & 0.5912±0.0089 & 0.5959±0.0103 & 0.6612±0.0026 & 0.7162±0.0029 & 0.6732±0.0059 & 0.6888±0.0083 & 0.6865±0.0055 & 0.7010±0.0073 & 0.6530±0.0425 & 0.6755±0.0542 \\
        \midrule
    NeuronPattern & Neuron & 0.6303±0.0017 & 0.6856±0.0038 & 0.6531±0.0009 & 0.6873±0.0049 & 0.6710±0.0030 & 0.6998±0.0070 & 0.6865±0.0014 & 0.7204±0.0060 & 0.6602±0.0242 & 0.6983±0.0160 \\
    \midrule
    \multicolumn{12}{c}{Llama-3.2-3B-Instruct} \\
    \midrule
    \multirow{2}[3]{*}{Method} & \#Shot & \multicolumn{2}{c}{5} & \multicolumn{2}{c}{10} & \multicolumn{2}{c}{20} & \multicolumn{2}{c}{30} & \multicolumn{2}{c}{Avg.} \\
\cmidrule{2-12}          & InfoType & Macro-F1 & Acc   & Macro-F1 & Acc   & Macro-F1 & Acc   & Macro-F1 & Acc   & Macro-F1 & Acc \\
        \midrule
    Random & N/A   & 0.5640±0.0094 & 0.5646±0.0100 & 0.5149±0.0093 & 0.5156±0.0089 & 0.5647±0.0130 & 0.5663±0.0142 & 0.5826±0.0166 & 0.5847±0.0181 & 0.5566±0.0291 & 0.5578±0.0296 \\
       \midrule
    Patron & S.+E. & 0.6863±0.0024 & 0.7467±0.0023 & 0.6128±0.0042 & 0.6483±0.0084 & 0.4230±0.0133 & 0.4322±0.0112 & 0.5549±0.0141 & 0.5616±0.0173 & 0.5693±0.1113 & 0.5972±0.1335 \\
    \midrule
    \multirow{2}[1]{*}{Entropy} & Highest & 0.5271±0.0079 & 0.5272±0.0079 & 0.5766±0.0084 & 0.5795±0.0097 & 0.3883±0.0136 & 0.4039±0.0109 & 0.5132±0.0205 & 0.5139±0.0199 & 0.5013±0.0801 & 0.5061±0.0738 \\
          & Diverse & 0.6363±0.0086 & 0.6445±0.0105 & 0.6607±0.0083 & 0.6729±0.0105 & 0.7040±0.0123 & 0.7288±0.0176 & 0.6277±0.0088 & 0.6376±0.0109 & 0.6572±0.0342 & 0.6709±0.0415 \\
        \midrule
    \multirow{3}[0]{*}{TypiClust} & S.    & 0.4187±0.0106 & 0.4291±0.0087 & 0.6040±0.0092 & 0.6109±0.0112 & 0.6578±0.0049 & 0.6891±0.0108 & 0.4977±0.0171 & 0.4980±0.0174 & 0.5445±0.1070 & 0.5568±0.1158 \\
          & L.    & 0.7060±0.0004 & 0.7506±0.0036 & 0.6657±0.0038 & 0.6904±0.0071 & 0.5039±0.0150 & 0.5046±0.0145 & 0.5478±0.0117 & 0.5497±0.0132 & 0.6059±0.0955 & 0.6238±0.1158 \\
          & S.+L. & 0.4947±0.0114 & 0.4965±0.0106 & 0.5653±0.0128 & 0.5663±0.0136 & 0.5255±0.0103 & 0.5256±0.0104 & 0.5616±0.0151 & 0.5620±0.0156 & 0.5368±0.0333 & 0.5376±0.0329 \\
        \midrule
    \multirow{3}[1]{*}{fast-votek} & S.    & 0.5152±0.0100 & 0.5153±0.0100 & 0.5318±0.0083 & 0.5329±0.0090 & 0.5418±0.0187 & 0.5435±0.0205 & 0.4396±0.0199 & 0.4455±0.0173 & 0.5071±0.0463 & 0.5093±0.0441 \\
          & L.    & 0.5964±0.0060 & 0.6028±0.0077 & 0.6391±0.0069 & 0.6488±0.0087 & 0.5550±0.0133 & 0.5551±0.0137 & 0.4009±0.0163 & 0.4147±0.0133 & 0.5479±0.1038 & 0.5554±0.1013 \\
          & S.+L. & 0.5348±0.0078 & 0.5350±0.0081 & 0.6116±0.0128 & 0.6177±0.0150 & 0.4673±0.0151 & 0.4696±0.0139 & 0.5495±0.0146 & 0.5501±0.0152 & 0.5408±0.0592 & 0.5431±0.0608 \\
    \midrule
    \multirow{2}[1]{*}{votek} & S.+E. & 0.5231±0.0102 & 0.5233±0.0105 & 0.4890±0.0091 & 0.4903±0.0085 & 0.5407±0.0133 & 0.5436±0.0155 & 0.5509±0.0095 & 0.5552±0.0113 & 0.5259±0.0271 & 0.5281±0.0285 \\
          & S.+E.+L. & 0.5393±0.0097 & 0.5394±0.0097 & 0.6698±0.0044 & 0.6908±0.0072 & 0.5423±0.0124 & 0.5425±0.0125 & 0.5368±0.0145 & 0.5370±0.0147 & 0.5721±0.0652 & 0.5774±0.0756 \\
        \midrule
    NeuronPattern & Neuron & 0.6340±0.0042 & 0.6470±0.0062 & 0.6197±0.0087 & 0.6305±0.0113 & 0.6203±0.0054 & 0.6340±0.0084 & 0.6218±0.0080 & 0.6304±0.0100 & 0.6240±0.0068 & 0.6355±0.0079 \\
    \bottomrule
    \end{tabular}%

}
  
  \caption{Experimental results with Llama series on Edu-Feedback.}
  \label{tab:addlabel}%
\end{table*}%

\begin{table*}[htbp]
  \centering

    \scalebox{.5}{

        \begin{tabular}{cccccccccccc}
    \toprule
    \multicolumn{12}{c}{Llama-3.1-8B-Instruct} \\
    \midrule
    \multirow{2}[3]{*}{Method} & \#Shot & \multicolumn{2}{c}{5} & \multicolumn{2}{c}{10} & \multicolumn{2}{c}{20} & \multicolumn{2}{c}{30} & \multicolumn{2}{c}{Avg.} \\
\cmidrule{2-12}          & InfoType & Macro-F1 & Acc   & Macro-F1 & Acc   & Macro-F1 & Acc   & Macro-F1 & Acc   & Macro-F1 & Acc \\
        \midrule
    Random & N/A   & 0.7084±0.0195 & 0.7020±0.0072 & 0.7998±0.0155 & 0.7827±0.0170 & 0.7698±0.0115 & 0.7160±0.0151 & 0.8311±0.0112 & 0.8553±0.0129 & 0.7773±0.0523 & 0.7640±0.0703 \\
        \midrule
    Patron & S.+E. & 0.7563±0.0237 & 0.7480±0.0151 & 0.7817±0.0063 & 0.8020±0.0035 & 0.8550±0.0153 & 0.8440±0.0120 & 0.8343±0.0067 & 0.8420±0.0020 & 0.8068±0.0457 & 0.8090±0.0450 \\
    \midrule
    \multirow{2}[1]{*}{Entropy} & Highest & 0.8000±0.0120 & 0.7793±0.0162 & 0.8231±0.0098 & 0.8287±0.0081 & 0.7798±0.0152 & 0.7733±0.0170 & 0.7554±0.0173 & 0.7567±0.0081 & 0.7896±0.0288 & 0.7845±0.0310 \\
          & Diverse & 0.7936±0.0150 & 0.7807±0.0189 & 0.8535±0.0098 & 0.8413±0.0081 & 0.8532±0.0079 & 0.8560±0.0080 & 0.8150±0.0088 & 0.8300±0.0231 & 0.8288±0.0296 & 0.8270±0.0327 \\
       \midrule
    \multirow{3}[0]{*}{TypiClust} & S.    & 0.7759±0.0105 & 0.7440±0.0087 & 0.8031±0.0124 & 0.8067±0.0070 & 0.7810±0.0098 & 0.7867±0.0046 & 0.8320±0.0100 & 0.8247±0.0115 & 0.7980±0.0255 & 0.7905±0.0347 \\
          & L.    & 0.8035±0.0119 & 0.7927±0.0110 & 0.7733±0.0076 & 0.7533±0.0160 & 0.8362±0.0037 & 0.8360±0.0053 & 0.8121±0.0021 & 0.8267±0.0050 & 0.8063±0.0260 & 0.8022±0.0375 \\
          & S.+L. & 0.7916±0.0053 & 0.7787±0.0070 & 0.8131±0.0024 & 0.8180±0.0035 & 0.7931±0.0144 & 0.7660±0.0156 & 0.8569±0.0034 & 0.8527±0.0058 & 0.8137±0.0304 & 0.8038±0.0394 \\
       \midrule
    \multirow{3}[1]{*}{fast-votek} & S.    & 0.7945±0.0075 & 0.7867±0.0095 & 0.8425±0.0012 & 0.8400±0.0020 & 0.8243±0.0161 & 0.8187±0.0153 & 0.8490±0.0138 & 0.8553±0.0110 & 0.8276±0.0244 & 0.8252±0.0297 \\
          & L.    & 0.7669±0.0083 & 0.7573±0.0099 & 0.8762±0.0116 & 0.8687±0.0099 & 0.8359±0.0087 & 0.8267±0.0175 & 0.8470±0.0068 & 0.8460±0.0087 & 0.8315±0.0463 & 0.8247±0.0481 \\
          & S.+L. & 0.7506±0.0093 & 0.7433±0.0127 & 0.7838±0.0289 & 0.8013±0.0129 & 0.8580±0.0011 & 0.8560±0.0020 & 0.8452±0.0099 & 0.8493±0.0046 & 0.8094±0.0509 & 0.8125±0.0521 \\
    \midrule
    \multirow{2}[1]{*}{votek} & S.+E. & 0.8281±0.0060 & 0.8133±0.0050 & 0.8193±0.0045 & 0.8127±0.0070 & 0.8199±0.0103 & 0.8113±0.0136 & 0.8602±0.0074 & 0.8520±0.0104 & 0.8319±0.0193 & 0.8223±0.0198 \\
          & S.+E.+L. & 0.7473±0.0118 & 0.7467±0.0130 & 0.8659±0.0076 & 0.8600±0.0035 & 0.8146±0.0211 & 0.7913±0.0291 & 0.8447±0.0155 & 0.8420±0.0223 & 0.8181±0.0517 & 0.8100±0.0513 \\
        \midrule
    NeuronPattern & Neuron & 0.8346±0.0097 & 0.8320±0.0122 & 0.7918±0.0224 & 0.7707±0.0181 & 0.8595±0.0121 & 0.8507±0.0155 & 0.8400±0.0122 & 0.8506±0.0123 & 0.8341±0.0301 & 0.8233±0.0359 \\
    \midrule
    \multicolumn{12}{c}{Llama-3.2-3B-Instruct} \\
    \midrule
    \multirow{2}[3]{*}{Method} & \#Shot & \multicolumn{2}{c}{5} & \multicolumn{2}{c}{10} & \multicolumn{2}{c}{20} & \multicolumn{2}{c}{30} & \multicolumn{2}{c}{Avg.} \\
\cmidrule{2-12}          & InfoType & Macro-F1 & Acc   & Macro-F1 & Acc   & Macro-F1 & Acc   & Macro-F1 & Acc   & Macro-F1 & Acc \\
        \midrule
    Random & N/A   & 0.6374±0.0066 & 0.6460±0.0080 & 0.7095±0.0197 & 0.7207±0.0046 & 0.7265±0.0284 & 0.6960±0.0111 & 0.7203±0.0167 & 0.7633±0.0153 & 0.6985±0.0413 & 0.7065±0.0490 \\
        \midrule
    Patron & S.+E. & 0.6884±0.0098 & 0.7107±0.0083 & 0.6839±0.0066 & 0.7320±0.0053 & 0.7899±0.0107 & 0.7787±0.0170 & 0.7472±0.0161 & 0.7727±0.0101 & 0.7273±0.0507 & 0.7485±0.0326 \\
    \midrule
    \multirow{2}[1]{*}{Entropy} & Highest & 0.6892±0.0055 & 0.6673±0.0042 & 0.6709±0.0146 & 0.6453±0.0153 & 0.6520±0.0229 & 0.6413±0.0050 & 0.6384±0.0048 & 0.5967±0.0070 & 0.6626±0.0222 & 0.6377±0.0296 \\
          & Diverse & 0.7645±0.0086 & 0.7467±0.0064 & 0.7888±0.0152 & 0.7813±0.0133 & 0.7028±0.0166 & 0.7687±0.0070 & 0.6884±0.0363 & 0.7380±0.0193 & 0.7361±0.0482 & 0.7587±0.0199 \\
        \midrule
    \multirow{3}[0]{*}{TypiClust} & S.    & 0.7392±0.0226 & 0.7327±0.0208 & 0.7398±0.0083 & 0.7320±0.0060 & 0.6829±0.0300 & 0.7107±0.0186 & 0.8030±0.0236 & 0.8007±0.0136 & 0.7412±0.0491 & 0.7440±0.0391 \\
          & L.    & 0.7457±0.0064 & 0.7413±0.0070 & 0.6794±0.0104 & 0.6427±0.0133 & 0.7278±0.0082 & 0.7833±0.0133 & 0.6964±0.0746 & 0.7940±0.0111 & 0.7123±0.0300 & 0.7403±0.0690 \\
          & S.+L. & 0.6918±0.0159 & 0.6793±0.0121 & 0.7647±0.0148 & 0.7867±0.0058 & 0.7728±0.0256 & 0.7573±0.0127 & 0.7843±0.0206 & 0.7780±0.0140 & 0.7534±0.0418 & 0.7503±0.0489 \\
        \midrule
    \multirow{3}[1]{*}{fast-votek} & S.    & 0.7386±0.0185 & 0.7567±0.0150 & 0.7966±0.0111 & 0.8033±0.0163 & 0.7455±0.0235 & 0.7327±0.0172 & 0.7603±0.0254 & 0.7520±0.0260 & 0.7602±0.0259 & 0.7612±0.0300 \\
          & L.    & 0.6910±0.0044 & 0.6820±0.0080 & 0.7826±0.0059 & 0.7780±0.0080 & 0.7019±0.0121 & 0.7340±0.0250 & 0.6947±0.0215 & 0.7627±0.0201 & 0.7176±0.0436 & 0.7392±0.0422 \\
          & S.+L. & 0.7755±0.0095 & 0.7873±0.0110 & 0.7138±0.0265 & 0.7647±0.0064 & 0.6976±0.0350 & 0.7567±0.0205 & 0.7184±0.0319 & 0.7927±0.0190 & 0.7263±0.0340 & 0.7753±0.0174 \\
    \midrule
    \multirow{2}[1]{*}{votek} & S.+E. & 0.7520±0.0031 & 0.7587±0.0129 & 0.7517±0.0107 & 0.7453±0.0167 & 0.8012±0.0219 & 0.7867±0.0181 & 0.7165±0.0108 & 0.7940±0.0131 & 0.7554±0.0348 & 0.7712±0.0230 \\
          & S.+E.+L. & 0.7576±0.0163 & 0.7607±0.0136 & 0.7845±0.0034 & 0.7793±0.0095 & 0.7233±0.0312 & 0.7567±0.0117 & 0.7629±0.0363 & 0.8087±0.0095 & 0.7571±0.0254 & 0.7763±0.0237 \\
        \midrule
    NeuronPattern & Neuron & 0.7335±0.0072 & 0.7147±0.0050 & 0.7193±0.0199 & 0.7587±0.0092 & 0.7845±0.0259 & 0.8207±0.0147 & 0.7781±0.0117 & 0.7493±0.0117 & 0.7538±0.0323 & 0.7608±0.0442 \\
    \bottomrule
    \end{tabular}%

    }
    
  \caption{Experimental results with Llama series on TREC.}
  \label{tab:addlabel}%

\end{table*}%

%% file: Qwen-full_results.tex

\begin{table*}[htbp]
  \centering
  \scalebox{.5}{
    \begin{tabular}{cccccccccccc}
    \toprule
    \multicolumn{12}{c}{Qwen3-8B} \\
    \midrule
    \multirow{2}[3]{*}{Method} & \#Shot & \multicolumn{2}{c}{5} & \multicolumn{2}{c}{10} & \multicolumn{2}{c}{20} & \multicolumn{2}{c}{30} & \multicolumn{2}{c}{Avg.} \\
\cmidrule{2-12}          & InfoType & Macro-F1 & Acc   & Macro-F1 & Acc   & Macro-F1 & Acc   & Macro-F1 & Acc   & Macro-F1 & Acc \\
        \midrule
    Random & N/A   & 0.2980±0.0301 & 0.2798±0.0247 & 0.4128±0.0070 & 0.3864±0.0080 & 0.4448±0.0043 & 0.4216±0.0056 & 0.4726±0.0031 & 0.4646±0.0051 & 0.4071±0.0767 & 0.3881±0.0790 \\
        \midrule
    Patron & S.+E. & 0.3373±0.0236 & 0.3128±0.0210 & 0.4406±0.0058 & 0.4170±0.0065 & 0.4634±0.0065 & 0.4459±0.0091 & 0.4906±0.0027 & 0.4867±0.0042 & 0.4330±0.0670 & 0.4156±0.0743 \\
    \midrule
    \multirow{2}[1]{*}{Entropy} & Highest & 0.2748±0.0283 & 0.2621±0.0212 & 0.4231±0.0126 & 0.3990±0.0149 & 0.4509±0.0047 & 0.4332±0.0065 & 0.4853±0.0035 & 0.4827±0.0053 & 0.4085±0.0927 & 0.3943±0.0946 \\
          & Diverse & 0.3355±0.0201 & 0.3093±0.0179 & 0.4005±0.0124 & 0.3724±0.0125 & 0.4477±0.0042 & 0.4260±0.0055 & 0.4725±0.0035 & 0.4611±0.0063 & 0.4141±0.0603 & 0.3922±0.0662 \\
        \midrule
    \multirow{3}[0]{*}{TypiClust} & S.    & 0.2980±0.0277 & 0.2790±0.0220 & 0.4201±0.0082 & 0.3951±0.0095 & 0.4613±0.0040 & 0.4437±0.0060 & 0.4800±0.0034 & 0.4726±0.0066 & 0.4149±0.0818 & 0.3976±0.0853 \\
          & L.    & 0.3135±0.0292 & 0.2923±0.0244 & 0.3860±0.0142 & 0.3578±0.0154 & 0.4449±0.0039 & 0.4225±0.0043 & 0.4688±0.0024 & 0.4542±0.0040 & 0.4033±0.0692 & 0.3817±0.0719 \\
          & S.+L. & 0.2813±0.0242 & 0.2656±0.0190 & 0.4039±0.0142 & 0.3758±0.0151 & 0.4480±0.0041 & 0.4266±0.0050 & 0.4763±0.0024 & 0.4658±0.0044 & 0.4024±0.0861 & 0.3834±0.0868 \\
        \midrule
    \multirow{3}[1]{*}{FastVoteK} & S.    & 0.2475±0.0262 & 0.2420±0.0185 & 0.3876±0.0153 & 0.3615±0.0150 & 0.4551±0.0044 & 0.4362±0.0062 & 0.4823±0.0039 & 0.4767±0.0059 & 0.3931±0.1049 & 0.3791±0.1031 \\
          & L.    & 0.2762±0.0326 & 0.2629±0.0258 & 0.3758±0.0164 & 0.3485±0.0153 & 0.4335±0.0055 & 0.4092±0.0062 & 0.4787±0.0046 & 0.4709±0.0064 & 0.3911±0.0874 & 0.3729±0.0887 \\
          & S.+L. & 0.3327±0.0255 & 0.3091±0.0222 & 0.4099±0.0124 & 0.3823±0.0135 & 0.4531±0.0034 & 0.4332±0.0043 & 0.4883±0.0023 & 0.4847±0.0038 & 0.4210±0.0670 & 0.4024±0.0749 \\
    \midrule
    \multirow{2}[1]{*}{VoteK} & S.+E. & 0.3452±0.0191 & 0.3189±0.0175 & 0.3754±0.0139 & 0.3481±0.0135 & 0.4534±0.0028 & 0.4320±0.0039 & 0.4716±0.0057 & 0.4632±0.0081 & 0.4114±0.0607 & 0.3905±0.0682 \\
          & S.+E.+L. & 0.3335±0.0251 & 0.3098±0.0220 & 0.4092±0.0128 & 0.3819±0.0137 & 0.4528±0.0032 & 0.4331±0.0044 & 0.4880±0.0031 & 0.4844±0.0044 & 0.4209±0.0666 & 0.4023±0.0745 \\
       \midrule
    NeuronPattern & Neuron & 0.3611±0.0253 & 0.3341±0.0232 & 0.4166±0.0102 & 0.3911±0.0117 & 0.4706±0.0035 & 0.4593±0.0054 & 0.4886±0.0023 & 0.4866±0.0036 & 0.4342±0.0575 & 0.4178±0.0687 \\
    \midrule
    \multicolumn{12}{c}{Qwen3-4B-Instruct-2507} \\
    \midrule
    \multirow{2}[3]{*}{Method} & \#Shot & \multicolumn{2}{c}{5} & \multicolumn{2}{c}{10} & \multicolumn{2}{c}{20} & \multicolumn{2}{c}{30} & \multicolumn{2}{c}{Avg.} \\
\cmidrule{2-12}          & InfoType & Macro-F1 & Acc   & Macro-F1 & Acc   & Macro-F1 & Acc   & Macro-F1 & Acc   & Macro-F1 & Acc \\
        \midrule
    Random & N/A   & 0.4242±0.0144 & 0.4102±0.0211 & 0.4252±0.0017 & 0.4154±0.0038 & 0.4384±0.0069 & 0.4284±0.0093 & 0.4135±0.0070 & 0.3922±0.0094 & 0.4253±0.0102 & 0.4115±0.0150 \\
        \midrule
    Patron & S.+E. & 0.4089±0.0147 & 0.3929±0.0203 & 0.4422±0.0034 & 0.4382±0.0054 & 0.4199±0.0115 & 0.4033±0.0158 & 0.3557±0.0186 & 0.3301±0.0179 & 0.4067±0.0367 & 0.3911±0.0451 \\
    \midrule
    \multirow{2}[1]{*}{Entropy} & Highest & 0.4195±0.0198 & 0.4039±0.0273 & 0.4488±0.0093 & 0.4443±0.0121 & 0.4280±0.0097 & 0.4153±0.0126 & 0.4580±0.0038 & 0.4542±0.0041 & 0.4386±0.0179 & 0.4295±0.0237 \\
          & Diverse & 0.4431±0.0094 & 0.4374±0.0135 & 0.4150±0.0085 & 0.3987±0.0121 & 0.4326±0.0076 & 0.4201±0.0117 & 0.4194±0.0112 & 0.3990±0.0148 & 0.4275±0.0128 & 0.4138±0.0187 \\
        \midrule
    \multirow{3}[0]{*}{TypiClust} & S.    & 0.4325±0.0111 & 0.4252±0.0170 & 0.4474±0.0067 & 0.4434±0.0086 & 0.4594±0.0013 & 0.4565±0.0018 & 0.4350±0.0092 & 0.4231±0.0124 & 0.4436±0.0124 & 0.4371±0.0158 \\
          & L.    & 0.4079±0.0081 & 0.3887±0.0119 & 0.4099±0.0062 & 0.3930±0.0079 & 0.4063±0.0038 & 0.3863±0.0055 & 0.3649±0.0237 & 0.3372±0.0231 & 0.3973±0.0216 & 0.3763±0.0262 \\
          & S.+L. & 0.4096±0.0118 & 0.3934±0.0171 & 0.3883±0.0118 & 0.3626±0.0133 & 0.4210±0.0077 & 0.4069±0.0097 & 0.4369±0.0127 & 0.4284±0.0160 & 0.4140±0.0204 & 0.3978±0.0275 \\
        \midrule
    \multirow{3}[1]{*}{FastVoteK} & S.    & 0.3990±0.0098 & 0.3775±0.0129 & 0.4229±0.0047 & 0.4109±0.0065 & 0.4054±0.0099 & 0.3842±0.0139 & 0.3980±0.0109 & 0.3730±0.0132 & 0.4063±0.0115 & 0.3864±0.0170 \\
          & L.    & 0.3862±0.0239 & 0.3633±0.0263 & 0.4328±0.0133 & 0.4245±0.0186 & 0.4221±0.0095 & 0.4096±0.0120 & 0.4218±0.0102 & 0.4058±0.0135 & 0.4157±0.0203 & 0.4008±0.0263 \\
          & S.+L. & 0.4265±0.0164 & 0.4148±0.0232 & 0.4179±0.0074 & 0.4037±0.0115 & 0.4401±0.0033 & 0.4346±0.0065 & 0.4373±0.0085 & 0.4270±0.0122 & 0.4304±0.0102 & 0.4200±0.0136 \\
    \midrule
    \multirow{2}[1]{*}{VoteK} & S.+E. & 0.4148±0.0093 & 0.4010±0.0144 & 0.4132±0.0056 & 0.4013±0.0071 & 0.4252±0.0079 & 0.4154±0.0106 & 0.4078±0.0123 & 0.3860±0.0146 & 0.4153±0.0073 & 0.4009±0.0120 \\
          & S.+E.+L. & 0.3886±0.0388 & 0.3693±0.0445 & 0.3816±0.0314 & 0.3579±0.0348 & 0.4061±0.0048 & 0.3846±0.0076 & 0.4050±0.0090 & 0.3823±0.0116 & 0.3953±0.0121 & 0.3735±0.0124 \\
        \midrule
    NeuronPattern & Neuron & 0.4566±0.0025 & 0.4550±0.0024 & 0.4454±0.0010 & 0.4389±0.0007 & 0.4540±0.0014 & 0.4495±0.0009 & 0.4645±0.0010 & 0.4625±0.0021 & 0.4551±0.0079 & 0.4515±0.0099 \\
    \bottomrule
    \end{tabular}%

  }
    \caption{Experimental results with Qwen3 series on MMLU-Pro.}
  \label{tab:addlabel}%
\end{table*}%

\begin{table*}[htbp]
  \centering

\scalebox{.5}{

    \begin{tabular}{cccccccccccc}
    \toprule
    \multicolumn{12}{c}{Qwen3-8B} \\
    \midrule
    \multirow{2}[3]{*}{Method} & \#Shot & \multicolumn{2}{c}{5} & \multicolumn{2}{c}{10} & \multicolumn{2}{c}{20} & \multicolumn{2}{c}{30} & \multicolumn{2}{c}{Avg.} \\
\cmidrule{2-12}          & InfoType & Macro-F1 & Acc   & Macro-F1 & Acc   & Macro-F1 & Acc   & Macro-F1 & Acc   & Macro-F1 & Acc \\
        \midrule
    Random & N/A   & 0.5097±0.0015 & 0.5103±0.0014 & 0.5779±0.0016 & 0.5788±0.0017 & 0.6691±0.0007 & 0.6970±0.0014 & 0.6898±0.0013 & 0.7122±0.0020 & 0.6116±0.0836 & 0.6246±0.0967 \\
        \midrule
    Patron & S.+E. & 0.6753±0.0004 & 0.7265±0.0007 & 0.6391±0.0016 & 0.7363±0.0004 & 0.6421±0.0012 & 0.6550±0.0017 & 0.6809±0.0009 & 0.7180±0.0017 & 0.6593±0.0218 & 0.7089±0.0367 \\
    \midrule
    \multirow{2}[1]{*}{Entropy} & Highest & 0.5057±0.0016 & 0.5065±0.0015 & 0.5273±0.0019 & 0.5274±0.0019 & 0.5726±0.0019 & 0.5732±0.0020 & 0.6158±0.0019 & 0.6205±0.0022 & 0.5554±0.0490 & 0.5569±0.0507 \\
          & Diverse & 0.6022±0.0019 & 0.6077±0.0022 & 0.6009±0.0017 & 0.6053±0.0020 & 0.6561±0.0004 & 0.6913±0.0017 & 0.6668±0.0018 & 0.7484±0.0003 & 0.6315±0.0349 & 0.6632±0.0695 \\
       \midrule
    \multirow{3}[0]{*}{TypiClust} & S.    & 0.4488±0.0017 & 0.4540±0.0015 & 0.6329±0.0016 & 0.6432±0.0020 & 0.6587±0.0014 & 0.6855±0.0023 & 0.6778±0.0008 & 0.7324±0.0009 & 0.6045±0.1055 & 0.6288±0.1221 \\
          & L.    & 0.6460±0.0016 & 0.6574±0.0019 & 0.6065±0.0016 & 0.6119±0.0018 & 0.6027±0.0032 & 0.6075±0.0036 & 0.6501±0.0014 & 0.6725±0.0024 & 0.6263±0.0252 & 0.6373±0.0325 \\
          & S.+L. & 0.4683±0.0019 & 0.4717±0.0017 & 0.5688±0.0018 & 0.5695±0.0019 & 0.6466±0.0010 & 0.6655±0.0017 & 0.6701±0.0011 & 0.6934±0.0018 & 0.5885±0.0911 & 0.6000±0.1006 \\
        \midrule
    \multirow{3}[1]{*}{FastVoteK} & S.    & 0.6011±0.0013 & 0.6083±0.0016 & 0.6460±0.0009 & 0.6723±0.0015 & 0.6660±0.0004 & 0.6983±0.0008 & 0.6454±0.0016 & 0.6669±0.0024 & 0.6396±0.0274 & 0.6615±0.0380 \\
          & L.    & 0.5339±0.0021 & 0.5346±0.0023 & 0.6383±0.0023 & 0.6526±0.0028 & 0.6612±0.0012 & 0.6787±0.0018 & 0.6489±0.0023 & 0.6622±0.0029 & 0.6206±0.0585 & 0.6320±0.0658 \\
          & S.+L. & 0.4712±0.0020 & 0.4742±0.0018 & 0.6555±0.0014 & 0.6826±0.0021 & 0.6421±0.0026 & 0.6519±0.0031 & 0.6879±0.0005 & 0.7150±0.0012 & 0.6142±0.0972 & 0.6310±0.1076 \\
    \midrule
    \multirow{2}[1]{*}{VoteK} & S.+E. & 0.6456±0.0009 & 0.6729±0.0016 & 0.6651±0.0006 & 0.6973±0.0012 & 0.6685±0.0007 & 0.7127±0.0013 & 0.6945±0.0011 & 0.7210±0.0020 & 0.6684±0.0201 & 0.7010±0.0211 \\
          & S.+E.+L. & 0.6401±0.0008 & 0.6607±0.0012 & 0.6717±0.0002 & 0.7085±0.0010 & 0.6107±0.0011 & 0.6188±0.0016 & 0.6008±0.0026 & 0.6035±0.0028 & 0.6308±0.0320 & 0.6479±0.0471 \\
        \midrule
    NeuronPattern & Neuron & 0.6679±0.0001 & 0.7088±0.0009 & 0.6680±0.0010 & 0.6928±0.0014 & 0.6724±0.0005 & 0.7033±0.0012 & 0.6424±0.0023 & 0.7378±0.0004 & 0.6627±0.0137 & 0.7107±0.0193 \\
    \midrule
    \multicolumn{12}{c}{Qwen3-4B-Instruct-2507} \\
    \midrule
    \multirow{2}[3]{*}{Method} & \#Shot & \multicolumn{2}{c}{5} & \multicolumn{2}{c}{10} & \multicolumn{2}{c}{20} & \multicolumn{2}{c}{30} & \multicolumn{2}{c}{Avg.} \\
\cmidrule{2-12}          & InfoType & Macro-F1 & Acc   & Macro-F1 & Acc   & Macro-F1 & Acc   & Macro-F1 & Acc   & Macro-F1 & Acc \\
        \midrule
    Random & N/A   & 0.5612±0.0016 & 0.5612±0.0017 & 0.6119±0.0027 & 0.6145±0.0029 & 0.7115±0.0008 & 0.7382±0.0013 & 0.7283±0.0009 & 0.7528±0.0015 & 0.6532±0.0800 & 0.6667±0.0937 \\
        \midrule
    Patron & S.+E. & 0.7022±0.0011 & 0.7660±0.0003 & 0.6624±0.0014 & 0.7554±0.0003 & 0.6616±0.0008 & 0.6980±0.0016 & 0.7142±0.0004 & 0.7487±0.0013 & 0.6851±0.0271 & 0.7420±0.0302 \\
    \midrule
    \multirow{2}[1]{*}{Entropy} & Highest & 0.4661±0.0023 & 0.4712±0.0021 & 0.5513±0.0019 & 0.5513±0.0019 & 0.5849±0.0023 & 0.5855±0.0024 & 0.6178±0.0016 & 0.6208±0.0018 & 0.5550±0.0652 & 0.5572±0.0640 \\
          & Diverse & 0.6596±0.0018 & 0.6675±0.0021 & 0.5964±0.0017 & 0.5978±0.0018 & 0.7141±0.0017 & 0.7344±0.0026 & 0.7151±0.0014 & 0.7371±0.0018 & 0.6713±0.0563 & 0.6842±0.0660 \\
        \midrule
    \multirow{3}[0]{*}{TypiClust} & S.    & 0.5132±0.0018 & 0.5133±0.0018 & 0.5820±0.0032 & 0.5831±0.0034 & 0.6955±0.0015 & 0.7149±0.0022 & 0.7129±0.0006 & 0.7486±0.0014 & 0.6259±0.0949 & 0.6400±0.1106 \\
          & L.    & 0.6381±0.0015 & 0.6433±0.0017 & 0.6535±0.0012 & 0.6614±0.0014 & 0.6163±0.0034 & 0.6193±0.0037 & 0.6373±0.0022 & 0.6441±0.0026 & 0.6363±0.0153 & 0.6420±0.0173 \\
          & S.+L. & 0.5047±0.0013 & 0.5055±0.0012 & 0.5915±0.0023 & 0.5937±0.0024 & 0.6528±0.0024 & 0.6656±0.0029 & 0.7037±0.0017 & 0.7283±0.0024 & 0.6132±0.0856 & 0.6233±0.0958 \\
        \midrule
    \multirow{3}[1]{*}{FastVoteK} & S.    & 0.6307±0.0021 & 0.6366±0.0024 & 0.6824±0.0008 & 0.7187±0.0003 & 0.6921±0.0009 & 0.7367±0.0009 & 0.6885±0.0006 & 0.7345±0.0013 & 0.6734±0.0287 & 0.7066±0.0474 \\
          & L.    & 0.5935±0.0015 & 0.5321±0.0013 & 0.6769±0.0008 & 0.7126±0.0024 & 0.6901±0.0013 & 0.7006±0.0017 & 0.6861±0.0006 & 0.7474±0.0012 & 0.6617±0.0457 & 0.6732±0.0961 \\
          & S.+L. & 0.5321±0.0013 & 0.5991±0.0018 & 0.6903±0.0018 & 0.6967±0.0012 & 0.6847±0.0011 & 0.7138±0.0017 & 0.7124±0.0003 & 0.7072±0.0011 & 0.6549±0.0827 & 0.6792±0.0539 \\
    \midrule
    \multirow{2}[1]{*}{VoteK} & S.+E. & 0.6003±0.0013 & 0.6022±0.0014 & 0.6658±0.0009 & 0.6797±0.0011 & 0.6943±0.0012 & 0.7203±0.0017 & 0.6934±0.0009 & 0.7224±0.0017 & 0.6635±0.0442 & 0.6812±0.0562 \\
          & S.+E.+L. & 0.6023±0.0012 & 0.6059±0.0013 & 0.6874±0.0006 & 0.7092±0.0011 & 0.7131±0.0007 & 0.7363±0.0011 & 0.7225±0.0003 & 0.7575±0.0006 & 0.6813±0.0548 & 0.7022±0.0672 \\
        \midrule
    NeuronPattern & Neuron & 0.7081±0.0002 & 0.7651±0.0002 & 0.7097±0.0004 & 0.7353±0.0006 & 0.6659±0.0009 & 0.6747±0.0011 & 0.6833±0.0001 & 0.7248±0.0008 & 0.6918±0.0211 & 0.7250±0.0376 \\
    \bottomrule
    \end{tabular}%

}

  \caption{Experimental results with Qwen3 series on Edu-Feedback.}

  \label{tab:addlabel}%
\end{table*}%

\begin{table*}[htbp]
  \centering

  \scalebox{.5}{

      \begin{tabular}{cccccccccccc}
    \toprule
    \multicolumn{12}{c}{Qwen3-8B} \\
    \midrule
    \multirow{2}[3]{*}{Method} & \#Shot & \multicolumn{2}{c}{5} & \multicolumn{2}{c}{10} & \multicolumn{2}{c}{20} & \multicolumn{2}{c}{30} & \multicolumn{2}{c}{Avg.} \\
\cmidrule{2-12}          & InfoType & Macro-F1 & Acc   & Macro-F1 & Acc   & Macro-F1 & Acc   & Macro-F1 & Acc   & Macro-F1 & Acc \\
        \midrule
    Random & N/A   & 0.8295±0.0023 & 0.8207±0.0031 & 0.8246±0.0071 & 0.7947±0.0095 & 0.8467±0.0004 & 0.8113±0.0012 & 0.8737±0.0027 & 0.8640±0.0040 & 0.8436±0.0222 & 0.8227±0.0296 \\
        \midrule
    Patron & S.+E. & 0.8189±0.0030 & 0.8213±0.0031 & 0.7989±0.0082 & 0.8160±0.0072 & 0.8259±0.0048 & 0.8093±0.0070 & 0.8543±0.0017 & 0.8473±0.0031 & 0.8245±0.0230 & 0.8235±0.0166 \\
    \midrule
    \multirow{2}[1]{*}{Entropy} & Highest & 0.8069±0.0099 & 0.8167±0.0042 & 0.8194±0.0013 & 0.8220±0.0020 & 0.8356±0.0045 & 0.8327±0.0061 & 0.8861±0.0037 & 0.8847±0.0042 & 0.8370±0.0348 & 0.8390±0.0312 \\
          & Diverse & 0.8532±0.0019 & 0.8473±0.0012 & 0.8579±0.0053 & 0.8413±0.0031 & 0.8699±0.0057 & 0.8567±0.0023 & 0.8410±0.0004 & 0.8180±0.0020 & 0.8555±0.0119 & 0.8408±0.0165 \\
       \midrule
    \multirow{3}[0]{*}{TypiClust} & S.    & 0.8003±0.0044 & 0.7980±0.0069 & 0.8282±0.0038 & 0.8240±0.0035 & 0.8507±0.0029 & 0.8460±0.0020 & 0.8436±0.0071 & 0.8433±0.0031 & 0.8307±0.0223 & 0.8278±0.0222 \\
          & L.    & 0.8212±0.0033 & 0.8353±0.0012 & 0.7995±0.0058 & 0.7673±0.0070 & 0.8570±0.0023 & 0.8493±0.0023 & 0.8637±0.0006 & 0.8527±0.0012 & 0.8354±0.0303 & 0.8262±0.0399 \\
          & S.+L. & 0.8413±0.0035 & 0.8333±0.0050 & 0.8467±0.0058 & 0.8380±0.0053 & 0.8562±0.0038 & 0.8287±0.0050 & 0.8588±0.0054 & 0.8400±0.0000 & 0.8508±0.0082 & 0.8350±0.0051 \\
        \midrule
    \multirow{3}[1]{*}{FastVoteK} & S.    & 0.8461±0.0047 & 0.8413±0.0064 & 0.8571±0.0012 & 0.8420±0.0020 & 0.8172±0.0051 & 0.7993±0.0070 & 0.8411±0.0005 & 0.8393±0.0012 & 0.8404±0.0168 & 0.8305±0.0208 \\
          & L.    & 0.8416±0.0011 & 0.8420±0.0020 & 0.8587±0.0051 & 0.8533±0.0012 & 0.8773±0.0006 & 0.8567±0.0012 & 0.8384±0.0021 & 0.8193±0.0050 & 0.8540±0.0179 & 0.8428±0.0169 \\
          & S.+L. & 0.8314±0.0018 & 0.8427±0.0012 & 0.8661±0.0030 & 0.8653±0.0012 & 0.8642±0.0024 & 0.8473±0.0023 & 0.8836±0.0028 & 0.8593±0.0023 & 0.8613±0.0218 & 0.8537±0.0105 \\
    \midrule
    \multirow{2}[1]{*}{VoteK} & S.+E. & 0.8392±0.0044 & 0.8307±0.0058 & 0.8411±0.0033 & 0.8320±0.0035 & 0.8458±0.0013 & 0.8387±0.0012 & 0.8309±0.0049 & 0.8527±0.0050 & 0.8393±0.0062 & 0.8385±0.0101 \\
          & S.+E.+L. & 0.8441±0.0016 & 0.8433±0.0023 & 0.8438±0.0008 & 0.8460±0.0000 & 0.8374±0.0032 & 0.8080±0.0035 & 0.8309±0.0027 & 0.8053±0.0061 & 0.8391±0.0062 & 0.8257±0.0220 \\
        \midrule
    NeuronPattern & Neuron & 0.8237±0.0040 & 0.8287±0.0031 & 0.8654±0.0009 & 0.8727±0.0012 & 0.8856±0.0015 & 0.8600±0.0020 & 0.8717±0.0056 & 0.8687±0.0023 & 0.8616±0.0266 & 0.8575±0.0199 \\
    \midrule
    \multicolumn{12}{c}{Qwen3-4B-Instruct-2507} \\
    \midrule
    \multirow{2}[3]{*}{Method} & \#Shot & \multicolumn{2}{c}{5} & \multicolumn{2}{c}{10} & \multicolumn{2}{c}{20} & \multicolumn{2}{c}{30} & \multicolumn{2}{c}{Avg.} \\
\cmidrule{2-12}          & InfoType & Macro-F1 & Acc   & Macro-F1 & Acc   & Macro-F1 & Acc   & Macro-F1 & Acc   & Macro-F1 & Acc \\
        \midrule
    Random & N/A   & 0.8039±0.0016 & 0.8047±0.0012 & 0.8754±0.0009 & 0.8713±0.0012 & 0.8848±0.0013 & 0.8820±0.0020 & 0.8698±0.0029 & 0.8813±0.0031 & 0.8585±0.0369 & 0.8598±0.0371 \\
        \midrule
    Patron & S.+E. & 0.8193±0.0015 & 0.8267±0.0012 & 0.8258±0.0039 & 0.8387±0.0012 & 0.8646±0.0021 & 0.8773±0.0031 & 0.8598±0.0048 & 0.8673±0.0012 & 0.8424±0.0231 & 0.8525±0.0238 \\
    \midrule
    \multirow{2}[1]{*}{Entropy} & Highest & 0.8405±0.0019 & 0.8393±0.0023 & 0.8521±0.0035 & 0.8540±0.0035 & 0.8592±0.0044 & 0.8620±0.0040 & 0.8681±0.0035 & 0.8713±0.0031 & 0.8550±0.0116 & 0.8567±0.0136 \\
          & Diverse & 0.8363±0.0015 & 0.8473±0.0023 & 0.8613±0.0012 & 0.8753±0.0012 & 0.8712±0.0024 & 0.8707±0.0023 & 0.8807±0.0010 & 0.8753±0.0012 & 0.8624±0.0191 & 0.8672±0.0134 \\
        \midrule
    \multirow{3}[0]{*}{TypiClust} & S.    & 0.8344±0.0014 & 0.8633±0.0031 & 0.8558±0.0036 & 0.8667±0.0031 & 0.8688±0.0024 & 0.8653±0.0012 & 0.8509±0.0083 & 0.8693±0.0023 & 0.8525±0.0142 & 0.8662±0.0025 \\
          & L.    & 0.8696±0.0008 & 0.8707±0.0012 & 0.8655±0.0062 & 0.8553±0.0023 & 0.8615±0.0015 & 0.8680±0.0020 & 0.8504±0.0060 & 0.8580±0.0020 & 0.8618±0.0083 & 0.8630±0.0075 \\
          & S.+L. & 0.8341±0.0035 & 0.8467±0.0012 & 0.8428±0.0009 & 0.8507±0.0012 & 0.8732±0.0008 & 0.8607±0.0012 & 0.8704±0.0009 & 0.8753±0.0012 & 0.8551±0.0196 & 0.8583±0.0128 \\
        \midrule
    \multirow{3}[1]{*}{FastVoteK} & S.    & 0.8677±0.0029 & 0.8660±0.0035 & 0.8690±0.0011 & 0.8687±0.0012 & 0.8656±0.0007 & 0.8653±0.0012 & 0.8878±0.0028 & 0.8827±0.0031 & 0.8725±0.0103 & 0.8707±0.0081 \\
          & L.    & 0.8260±0.0000 & 0.8520±0.0000 & 0.8556±0.0010 & 0.8533±0.0012 & 0.8743±0.0015 & 0.8780±0.0020 & 0.8882±0.0017 & 0.8847±0.0023 & 0.8610±0.0269 & 0.8670±0.0168 \\
          & S.+L. & 0.8151±0.0049 & 0.8213±0.0023 & 0.8559±0.0005 & 0.8607±0.0012 & 0.8777±0.0004 & 0.8807±0.0012 & 0.8402±0.0013 & 0.8633±0.0012 & 0.8472±0.0263 & 0.8565±0.0251 \\
    \midrule
    \multirow{2}[1]{*}{VoteK} & S.+E. & 0.8442±0.0040 & 0.8380±0.0053 & 0.8515±0.0000 & 0.8480±0.0000 & 0.8551±0.0001 & 0.8660±0.0000 & 0.8608±0.0007 & 0.8487±0.0012 & 0.8529±0.0069 & 0.8502±0.0116 \\
          & S.+E.+L. & 0.8002±0.0043 & 0.8180±0.0020 & 0.8747±0.0035 & 0.8747±0.0012 & 0.8587±0.0061 & 0.8473±0.0023 & 0.8804±0.0032 & 0.8453±0.0046 & 0.8535±0.0367 & 0.8463±0.0231 \\
        \midrule
    NeuronPattern & Neuron & 0.8390±0.0025 & 0.8613±0.0012 & 0.8624±0.0022 & 0.8713±0.0031 & 0.8771±0.0000 & 0.8860±0.0000 & 0.8800±0.0038 & 0.8947±0.0042 &  0.8646±0.0188 & 0.8783±0.0149 \\
    \bottomrule
    \end{tabular}%

  }

    \caption{Experimental results with Qwen3 series on TREC.}

  \label{tab:qwen-trec}%
\end{table*}%